%% file: main.tex
\documentclass[10pt]{article} 
\usepackage[accepted]{tmlr}

\input{math_commands.tex}

\usepackage{arydshln}
\usepackage{adjustbox}
\usepackage{amsmath}
\usepackage{makecell}
\usepackage{pifont}
\usepackage{url}            
\usepackage{booktabs}       
\usepackage{colortbl}       
\usepackage{amsfonts}       
\usepackage{nicefrac}       
\usepackage{microtype}      
\usepackage{xcolor}
\usepackage{enumitem}
\usepackage{wrapfig}
\usepackage{algorithm}
\usepackage{multirow}
\usepackage{nccmath}
\usepackage{algpseudocode}
\usepackage{xspace}
\newcommand{\ie}{i.e.\xspace}

\newcommand{\rev}[1]{#1}

\title{Adapting Vision Foundation Models with Cascaded Semantics}

\author{\name Xi Xiao \\
      \addr University of Alabama at Birmingham
      \AND
      \name Xingjian Li \\
      \addr Carnegie Mellon University
      \AND
      \name Cheng Han\thanks{Co-advising} \\
      \addr University of Missouri--Kansas City
      \AND
      \name Tianyang Wang\footnotemark[1] \\
      \addr University of Alabama at Birmingham
      \AND
      \name Lin Zhao \\
      \addr Northeastern University
      \AND
      \name Yunbei Zhang \\
      \addr Tulane University
      \AND
      \name Guosheng Hu \\
      \addr University of Bristol
      \AND
      \name Runmin Jiang \\
      \addr Carnegie Mellon University
      \AND
      \name Xi Li \\
      \addr University of Alabama at Birmingham
      \AND
      \name Xiao Wang \\
      \addr Oak Ridge National Laboratory
      \AND
      \name Min Xu \\
      \addr Carnegie Mellon University}

\definecolor{naturegreen}{RGB}{0,102,85}
\definecolor{cvprblue}{rgb}{0.21,0.49,0.74}
\definecolor{lightblue}{rgb}{0.796, 0.894, 0.9808}
\usepackage[pagebackref,breaklinks,colorlinks,allcolors=cvprblue]{hyperref}

\def\month{08}
\def\year{2026}
\def\openreview{\url{https://openreview.net/forum?id=SSsobNZJPO}}

\begin{document}

\maketitle

\begin{abstract}
Prompt tuning, a leading parameter-efficient adaptation paradigm in NLP, has recently been extended to computer vision. Visual prompt tuning (VPT) adapts pre-trained vision transformers (ViTs) by updating a small set of additional prompt parameters. However, existing visual prompts are randomly initialized and do not exploit prior knowledge, such as instructions in NLP. We address this gap by injecting two complementary semantic priors into VPT. Fundamental image priors, including color, texture, and shape, are extracted with classical hand-crafted operators and injected into the input space, while self-attention maps provide instance-aware semantics in the feature space. We further propose a cascaded scheme that integrates both priors throughout ViT adaptation. Experiments on 34 challenging image classification datasets demonstrate superior downstream adaptation while tuning only 0.74\% of ViT parameters. Project page: \url{https://xixiaouab.github.io/Cascaded-Semantics/}.
\end{abstract}

\input{sec/1_intro}

\input{sec/3_method}

\input{sec/4_experiment}

\input{sec/5_conclusion}

\section*{Acknowledgments}
This manuscript was co-authored by Oak Ridge National Laboratory (ORNL), operated by UT-Battelle, LLC under Contract No. DE-AC05-00OR22725 with the U.S. Department of Energy. Any subjective views or opinions expressed in this paper do not necessarily represent those of the U.S. Department of Energy or the United States Government. This research was supported by the National Science Foundation under Grant No. 2450068. This work was supported in part by U.S. NSF grants DBI-2238093, DBI-2422619, IIS-2211597, and MCB-2205148. This work was also supported in part by an Amazon Research Award (Fall 2025 CFP).

\bibliography{main}
\bibliographystyle{tmlr}

\clearpage
\input{sec/X_suppl}

\end{document}

%% file: math_commands.tex
\usepackage{amsmath,amsfonts,bm}

\def\eqref#1{equation~\ref{#1}}

\def\1{\bm{1}}

\DeclareMathAlphabet{\mathsfit}{\encodingdefault}{\sfdefault}{m}{sl}
\SetMathAlphabet{\mathsfit}{bold}{\encodingdefault}{\sfdefault}{bx}{n}



%% file: sec/1_intro.tex
\section{Introduction}

Adapting pre-trained large-scale vision models to downstream tasks through parameter-efficient fine-tuning (PEFT) \citep{houlsby2019parameter,li2021prefix,hu2022lora,xiao2026structlora,wang2026ctrlora} has been shown to be practical, especially when the downstream data is limited. Visual prompt tuning (VPT) \citep{jia2022visual,xiao2025promptbased} is one of the most famous PEFT methods that does not resort to changing the structure or parameters of pre-trained vision transformers (ViTs) \citep{dosovitskiy2020image}, leading to a highly convenient pipeline. Unlike other methods, such as \citep{rebuffi2017learning,hu2022lora} that update pre-trained parameters, and \citep{chen2022adaptformer} that restructures the transformer block, VPT incorporates a small amount of learnable parameters (\ie, prompt) into the input and feature spaces of a ViT, and only updates them during fine-tuning with gradient descent. Such a paradigm yields surprisingly promising results in various scenarios, even exceeding the full fine-tuning that was deemed the most reliable tuning fashion~\citep{houlsby2019parameter,han2024facing,xiao2025visual,xiao2026layer,li2026pib}.

Though highly successful, \textit{\textbf{two critical challenges}} naturally arise in current VPT research. 
\textbf{I.~Random prompt initialization.} The first problem lies in the initialization of these visual learnable prompts: 
Current visual prompts are generally designed to be randomly initialized without considering any prior knowledge. In NLP, very differently, 
prompts are typically served as explicit instructions or context added to input text, guiding model’s behavior based on previously seen textual information \citep{brown2020language, petroni2019language}. The absence of such a prior in the visual domain makes
visual prompt more akin to black-box parameter optimization (\ie, tuning prompt) \citep{bahng2022visual}, which could potentially lead to 
sub-optimal performance. 
\textbf{II.~\rev{Prompt behavior is rarely analyzed at the representation level}.}
In NLP, both hard (not learnable)~\citep{brown2020language,petroni2019language,Schick2020ExploitingCF} and soft (learnable)~\citep{li2021prefix,lester2021power} prompts can be applied simultaneously. Although the soft ones may not be directly meaningful to humans, the hard ones are usually given by humans and are thus naturally interpretable. In VPT design, as these learnable prompts are directly updated via gradient descent, leaving little room for human interpretation \citep{bahng2022visual}. Some research tries to involve intrepretability via attention distribution~\citep{zeng2024visual} or \textit{post-hoc} explanation~\citep{wang2022visual} (e.g, GradCAM~\citep{zeng2024visual,han2023e2vpt,han2024facing}). However, they stop at explaining visual prompts as a whole optimization objective after training, ignoring the characteristics of per-input variants or case-by-case instance awareness~\citep{xiao2025visual,xiao2025vapt,li2026pib}.
\begin{wrapfigure}{r}{0.45\textwidth}
    \centering
    \includegraphics[width=\linewidth]{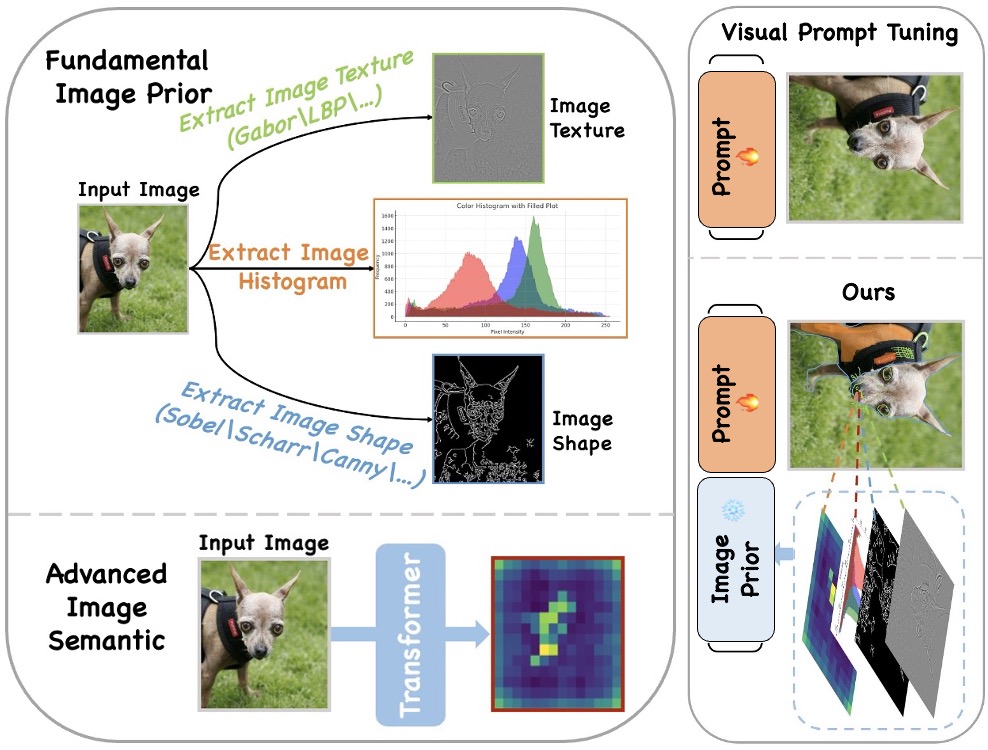}
    \caption{Overview of cascaded semantic prompting. Fundamental image priors (\ie, color, texture, and shape) condition the input space, while instance-aware self-attention semantics guide intermediate features.}
    \label{fig:image_priors}
    \vspace{-0.5em}
\end{wrapfigure}

Motivated by this, we explore incorporating interpretable prior knowledge during training as prompt into the vision tuning, which kills two birds with one stone: 
\ding{172} \rev{We do not remove the randomized learnable prompt; we anchor it. The fixed prior tokens (color, texture, shape, and the cascaded self-attention map) provide a structured warm start, so the random tokens are now optimized around \textit{semantically meaningful directions} rather than from scratch. This addresses \textit{\textbf{challenge I}}, while keeping the learnability that prompt tuning relies on.}
\ding{173} Since the fundamental image prior comes from human-understandable information (e.g, color/texture/shape), \rev{its effect on the learned features can be checked with common post-hoc tools (cosine similarity, IoU, GradCAM, t-SNE, and mutual information), which we report in Sec.~\ref{sec:rep_analysis} and Appendix Sec.~\ref{sec:cosine_analysis}. This gives an \textit{instance-aware, image-grounded} adaptation path}~\citep{wang2022visual, biehl2016prototype,swain1991color,manjunath1996texture,Dalal2005HistogramsOO} \rev{that is more transparent than random prompts}, \rev{which addresses} \textit{\textbf{challenge II}}.

Specifically, we consider prompt positions in both input and feature spaces of a ViT \citep{jia2022visual} by injecting different types of priors. For \textit{\textbf{input space}}, ideally, the prior should meet three criteria: a) it contains a certain level of image semantics, thus interpretable to humans; b) it is free of learning, thus bringing no extra burden to the tuning; c) it can be easily pinned to input space. We thus leverage fundamental image priors, including color histograms \citep{swain1992indexing}, textures \citep{cimpoi2014describing}, and shapes \citep{zhang2004review}, as the prompt in input space. The motivation is that these semantics are well-known for their ability to reflect essential image clues and can be obtained by simply using classical, hand-crafted operators (e.g, Sobel \citep{kanopoulos1988design}) rather than learning \citep{swain1991color,manjunath1996texture,Dalal2005HistogramsOO} (see Sec. \S\ref{sec3-2}).
For \textit{\textbf{feature space}}, the ideal prior is expected to: a) inject information critical to the model's decision; b) be based on the knowledge learned from preceding layers so it can facilitate the tuning of subsequent layers. To meet these requirements, we leverage the instance-aware, case-by-case self-attention map \citep{parmar2018image,chefer2021transformer,han2022survey,khan2022transformers} as the injected prompt in feature space. The rationale lies in the fact that such a map well reflects per-image class activation semantics \citep{zhou2016learning}, and can be conveniently computed based on the features provided by transformer layers (see Sec. \S\ref{sec3-3}).

The following question turns into how to properly fuse the semantic prompts at different locations. We propose a simple yet effective scheme to cascade them. Specifically, for the prompt used in a specific transformer layer, it is formed by fusing the prompt and self-attention map in the preceding layer with that in the current layer, using skip connections \citep{huang2017densely, srivastava2015training, he2016deep, oyedotun2022everyone}. As such, the prompt in the current layer is nourished by the semantics learned from the preceding layers. Besides the semantic priors injected into the prompts, randomized learnable parameters are also utilized as part of the prompts, enabling gradient updates. A subsequent re-weighting adapter (see Sec. \S\ref{subsec:re-weight adapter}), demonstrated to be effective in \citep{kirichenko2022last}, is employed to enable flexible feature adaptation prior to the task head.

\noindent\textbf{Contributions.}
We (i) introduce fundamental image priors and instance-aware self-attention semantics as complementary visual prompts; (ii) develop a cascaded fusion design with a re-weighting adapter; and (iii) evaluate the method across 34 datasets using multiple backbones, controlled ablations, representation-level analyses, and end-to-end efficiency measurements.

\noindent\textbf{Claims and supporting evidence.}
The main empirical claims, stated separately from the contributions, are:
\begin{enumerate}[label=\textbf{C\arabic*:}, leftmargin=3.0em, itemsep=0pt, topsep=0.2em]
    \item \textbf{Accuracy and transfer.} With 0.74\% tuned parameters, the method improves aggregate accuracy over the compared PEFT baselines on FGVC, HTA, and VTAB-1k and transfers from ViT to Swin. \textbf{Evidence:} Tables~\ref{tab:vit} and~\ref{tab:performance_vtab1k}; three-seed results in Sec.~\ref{sec:std}.
    \item \textbf{Component effects.} Controlled removals support the contributions of fundamental priors, self-attention semantics, skip connections, and re-weighting. \textbf{Evidence:} Table~\ref{tab:ablation_fused} and Fig.~\ref{fig:ablations_four}.
    \item \textbf{Representation quality.} Semantic prompts improve feature--region alignment, localization, class separation, and label correlation; these measurements do not constitute a complete mechanistic explanation. \textbf{Evidence:} Sec.~\ref{sec:rep_analysis}, Figs.~\ref{fig:grad}--\ref{fig:mutual_information_comparison}, Table~\ref{tab:iou_performance}, and Appendix Fig.~\ref{fig:cosine}.
    \item \textbf{Semantic versus text prompts.} Under the evaluated CoOp/MaPLe base-to-novel protocol, semantic prompts slightly improve Base, Novel, and harmonic-mean accuracy over text-conditioned prompts. \textbf{Evidence:} Table~\ref{tab:performance}.
    \item \textbf{End-to-end efficiency.} Training cost and memory remain close to VPT, with inference and prior-extraction overhead measured explicitly. \textbf{Evidence:} Sec.~\ref{sec:efficiency_main}, Table~\ref{tab:efficiency_main}, and Sec.~\ref{sec:efficiency}.
\end{enumerate}

%% file: sec/3_method.tex
\section{Methodology}
\label{sec:methodology}

\subsection{Problem Definition}

Given a pre-trained large-scale vision model (e.g, ViT \citep{dosovitskiy2020image}) consisting of $N$ transformer layers, we use $\textcolor{blue}{f_i(\cdot)}$ to denote the feed-forwarding operation of the layer $i$, where $i\in {1,2,…,N}$. Rather than updating the model’s parameters, classical VPT \citep{jia2022visual} prepends a small amount of learnable parameters, namely $\textcolor{orange}{P}$, to input $\textcolor{blue}{X}$ and its features across layers (\ie, \textit{VPT-Deep}). As such, the first layer’s output can be modeled as $\textcolor{blue}{Y_1} = \textcolor{blue}{f_1}(\textcolor{orange}{P_1} \otimes \textcolor{blue}{X})$, while the $i^{th}$ layer’s output as $\textcolor{blue}{Y_i} = \textcolor{blue}{f_i}(\textcolor{orange}{P_i} \otimes \textcolor{blue}{Y_{i-1}})$, where $\otimes$ denotes concatenation. If~ $i=N$, then $\textcolor{blue}{Y_i}$ will be fed into a classification head $\textcolor{orange}{h}$, yielding logits as $\textcolor{orange}{h}(\textcolor{blue}{Y_i})$. During fine-tuning, only $\textcolor{orange}{P}$ and $\textcolor{orange}{h}$ are learnable by gradient descent. A special case of VPT, namely \textit{VPT-Shallow}, only prepends $\textcolor{orange}{P_1}$ to $\textcolor{blue}{X}$, while waiving all the other $\textcolor{orange}{P_i}$s ($i\in {2,…,N}$) in feature space. It has been found that the deep version outperforms the shallow one, motivating us to develop our new method on top of \textit{VPT-Deep}. The colors
\raisebox{0.1ex}{\textcolor{orange}{\rule{0.5em}{0.5em}}} and \raisebox{0.1ex}{\textcolor{blue}{\rule{0.5em}{0.5em}}} indicate \textcolor{orange}{trainable} and \textcolor{blue}{frozen} parameters, respectively.

\subsection{Visual Prior as Prompt for Input}
\label{sec3-2}

In the classical VPT, all the prompts are randomly initialized. Though general, this brings no possibility of customizing or adjusting these prompts case-by-case, image-by-image. The loss of uniqueness brings performance degradation and non-transparent operations.
Inspired by the hard prompts (unlearnable) in NLP, we aim to incorporate appropriate, instance-aware semantics to ViT in the form of hard prompt to solve these drawbacks. \rev{We clarify the terminology up front: throughout this paper, a \emph{hard prompt} means a \emph{fixed, non-learnable prompt component computed by a deterministic operator from the input image}. This is the sense used in early NLP prompting work, where hand-written instructions or templates serve as a fixed context~\citep{brown2020language,petroni2019language}. Our usage differs from recent NLP work that also calls discrete tokens ``hard prompts'' but \emph{learns} them through gradient-based or RL-based discrete optimization~\citep{wen2023hard,choi2024hard}: there, the tokens are themselves the optimization target; in our work, the hard prompt is given by a fixed operator (color histogram, Gabor, Sobel) and is never updated during fine-tuning.} As prompts can be prepended to both input and feature spaces, we develop two schemes of injecting semantics for both spaces, respectively. Specifically, for \textit{input space}, we leverage well-known hand-crafted operators to extract \textit{color histogram}, \textit{texture}, and \textit{shape} \rev{from the input image $X$ itself (not from any extracted feature)}, as fundamental image priors, denoted by $\textcolor{blue}{\sigma_c(X)}$, $\textcolor{blue}{\sigma_t(X)}$, and $\textcolor{blue}{\sigma_s(X)}$ respectively (e.g, $\textcolor{blue}{\sigma_t}$ as Gabor \citep{manjunath1996texture} and $\textcolor{blue}{\sigma_s}$ as Sobel \citep{kanopoulos1988design})  (see Sec. \S\ref{subsec:fipo} for more analysis). Then, we use these priors as hard prompts (non-learnable), concatenated with the randomized learnable prompt $\textcolor{orange}{P_1}$ in input space to form an overall prompt as 
\begin{equation}
    \textcolor{orange}{\tilde{P_1}} = \textcolor{orange}{P_1} \otimes \textcolor{orange}{FC}(\textcolor{blue}{\sigma_c(X)} \otimes \textcolor{blue}{\sigma_t(X)} \otimes \textcolor{blue}{\sigma_s(X)}),
\end{equation}
where a learnable Linear layer (\ie, $\textcolor{orange}{FC}$) is employed to adjust dimension as shown in Fig \ref{fig:main_architecture}(a). As a result, the first transformer layer's output becomes 
\begin{equation}
    \textcolor{blue}{Y_1} = \textcolor{blue}{f_1}(\textcolor{orange}{\tilde{P_1}} \otimes \textcolor{blue}{X}).
\end{equation}
For \textit{feature space}, we conjecture that the fundamental image priors are likely to be sub-optimal options since they are directly from input rather than feature. Therefore, we compute self-attention map as the visual semantics in feature space. However, how to properly utilize such semantics as prompt remains unclear, introduced next.

\subsection{Visual Semantics as Prompt for Features}\label{sec3-3}

\rev{Having handled the input space, we now turn to the feature space. The question is how to inject instance-aware semantics into the prompts at this level.} To answer this, we use self-attention maps as semantically rich, instance-aware signals that guide the prompts. As illustrated in Fig.~\ref{fig:main_architecture}(c), for layer $i$, where $i\in \{2,...,N\}$, we compute its self-attention map $\textcolor{blue}{\mathcal{A}_i}$ based on its output feature, and concatenate it with $\textcolor{orange}{P_i}$ as $\textcolor{orange}{FC}(\textcolor{blue}{\mathcal{A}_i})\otimes \textcolor{orange}{P_i}$, where a learnable $\textcolor{orange}{FC}$ layer is used to adjust dimension. \rev{To let semantics from the previous layer flow into the current one, we also concatenate this with the preceding prompt $\textcolor{orange}{P_{i-1}}$ and self-attention map $\textcolor{blue}{\mathcal{A}_{i-1}}$.} The overall prompt is
\begin{equation}
    \textcolor{orange}{\tilde{P_i}} = \textcolor{orange}{FC_i}(\textcolor{blue}{\mathcal{A}_i})\otimes \textcolor{orange}{P_i} \otimes \textcolor{orange}{FC_{i-1}}(\textcolor{blue}{\mathcal{A}_{i-1}})\otimes \textcolor{orange}{P_{i-1}},
\end{equation}
where both $\textcolor{orange}{P_i}$ and $\textcolor{orange}{P_{i-1}}$ are randomized learnable prompts. \rev{The second pair $(\textcolor{orange}{P_{i-1}}, \textcolor{blue}{\mathcal{A}_{i-1}})$ implements the \emph{skip-connection}: it directly carries the prompt and attention information from layer $i{-}1$ into layer $i$, analogous to a residual link~\citep{he2016deep,huang2017densely}. This way, the prompt at layer $i$ is anchored by the semantics already accumulated at layer $i{-}1$, instead of being computed from $\textcolor{blue}{\mathcal{A}_i}$ alone. Gradients are back-propagated through this skip path as well.}
Then, the input to the layer $i+1$ can be written as $\textcolor{orange}{\tilde{P_i}}\otimes \textcolor{blue}{Y_i}$, where $\textcolor{blue}{Y_i}$ denotes the output feature of the layer $i$. This fashion of prompting does not apply to the first layer, in which we use the prompt introduced in Sec. \S\ref{sec3-2}. Notably, during fine-tuning, only the learnable prompts $\textcolor{orange}{P_i}$s, where $i\in \{1,...,N\}$, and the $\textcolor{orange}{FC}$ layers are updated, without incurring much extra computing burden, thus maintaining the PEFT nature.

\subsection{Re-Weighting Adapter as the Final Puzzle}\label{subsec:re-weight adapter}
\begin{figure*}[!t]
    \centering
    \includegraphics[width=0.98\textwidth]{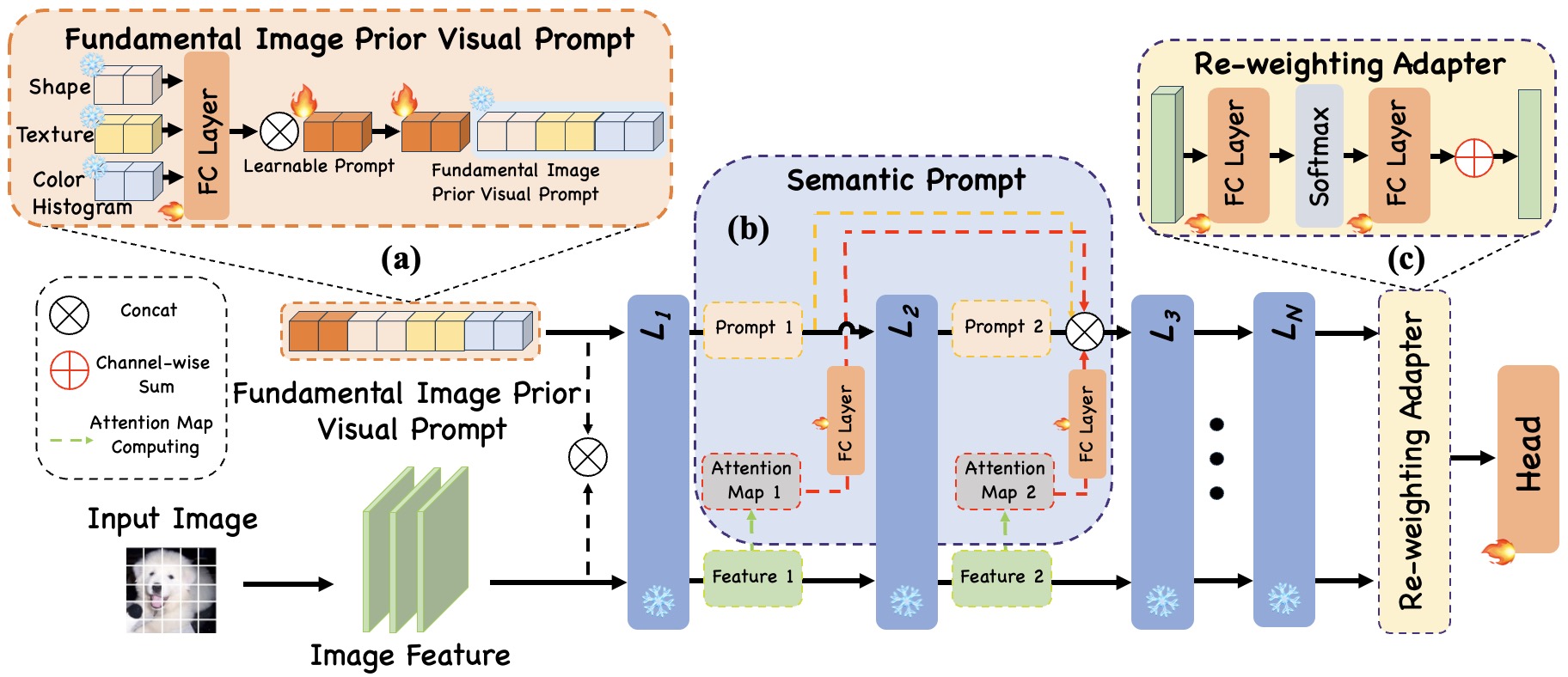}
    \caption{Overall architecture of our method.
    \textbf{(a) Fundamental Image Prior Visual Prompt:} hand-crafted priors (color histogram, texture, shape) are computed \rev{directly from the input image $X$ (i.e., in the input space, not from any deep feature)}, concatenated with a randomized learnable prompt, and projected by a lightweight FC for token-size alignment. \rev{The block labeled ``Image Feature'' in panel (a) denotes the resulting token embedding after this FC projection.}
    \textbf{(b) Re-Weighting Adapter:} a light two-layer linear module produces channel-wise weights to re-calibrate the final features before the head.
    \textbf{(c) Cascaded prompting in the backbone:} prompts are injected at selected layers $\{L_i\}$; the self-attention map from the previous layer is encoded and \emph{cascaded} forward as a semantic prompt, forming a skip-cascade that integrates fundamental priors with advanced image semantics.}
    \label{fig:main_architecture}
    \vspace{-1.3em}
\end{figure*}

Our scope is not solely limited to prompt tuning engineering questions; instead, we are inspired by the influence function in classical statistics that properly re-weighting and/or perturbing data or features can lead to improved generalization of deep models \citep{koh2017understanding}. This explains why fine-tuning works, as tuning the last linear layer(s) can be considered as re-weighting the features learned by a pre-trained model \citep{kirichenko2022last}, while tuning all the layers as perturbing the features. In our method, if prompt tuning is treated as being used for perturbing features, then it still needs an equivalent operation for feature re-weighting. Motivated by this, we propose a simple but effective re-weighting adapter, as shown in Fig.~\ref{fig:main_architecture}(b). Here, the output feature of $\textcolor{blue}{L_N}$ is fed into a combination of \textcolor{orange}{\textit{`FC-Softmax-FC'}}, whose output is channel-wisely summed and then fed into the learnable classification head $\textcolor{orange}{h}$.

\subsection{Why Don't Leave Visual Prior to Learning?}
In Sec. \S\ref{sec3-2}, we propose to inject the visual prior as prompt. Here, it is natural to raise a question: why don't we ask the model to learn such a prior automatically? In fact, if the model is trained from scratch, such prior could be better captured \citep{geirhos2018imagenet}. However, in the fine-tuning context, the model was pre-trained on source data that is different from target data, directly using the pre-trained model, which is frozen, might fail to effectively capture the fundamental image prior from target data \citep{BenDavid2010ATO,Torralba2011UnbiasedLA,Kornblith2018DoBI}. Therefore, it is reasonable to capture the prior with simple hand-crafted operators \citep{swain1991color,manjunath1996texture,Dalal2005HistogramsOO}, and then inject it as prompt to input space \citep{Touvron2020TrainingDI}.

\vspace{-0.1em}
\subsection{Are More Parameters Beneficial?}
In addition to the randomized learnable prompt, FC layers are also used for adjusting dimension, incorporating a few more learnable parameters. Then, it is necessary to investigate whether our method benefits from additional parameters. 
\textbf{\textit{Counter-intuitively, we observe that more parameters hurt the performance.}} We conduct the following studies, summarized in Fig.~\ref{fig:ablations_four}.
We replace the single FC layers in Fig.~\ref{fig:main_architecture} with deeper MLPs, involving more parameters; as shown in Fig.~\ref{fig:ablations_four}(a), this change brings a negative impact as depth increases.
We lengthen the randomized learnable prompts to include more parameters \textbf{(which has far exceeded the amount of parameters in our method)}, however, the performance is not \emph{monotonically} improving, as shown in Fig.~\ref{fig:ablations_four}(b), aligning with the observation in~\citep{jia2022visual}. Moving prompts across depths shows that targeted placement is better than indiscriminate/all-layer injection, even though the latter uses more tokens/parameters; see Fig.~\ref{fig:ablations_four}(c). Increasing the hidden dimension of the re-weighting/adapter brings only marginal gains and quickly saturates; see Fig.~\ref{fig:ablations_four}(d). All experiments are conducted exclusively, suggesting that simply adding more parameters is not beneficial \citep{Belkin2018ReconcilingMM,Nakkiran2019DeepDD, han2024facing}; where and how to use them matters more \citep{wang2024revisiting,chen2022adaptformer}.

%% file: sec/4_experiment.tex
\begin{figure*}[t]
    \centering
    \includegraphics[width=0.99\linewidth]{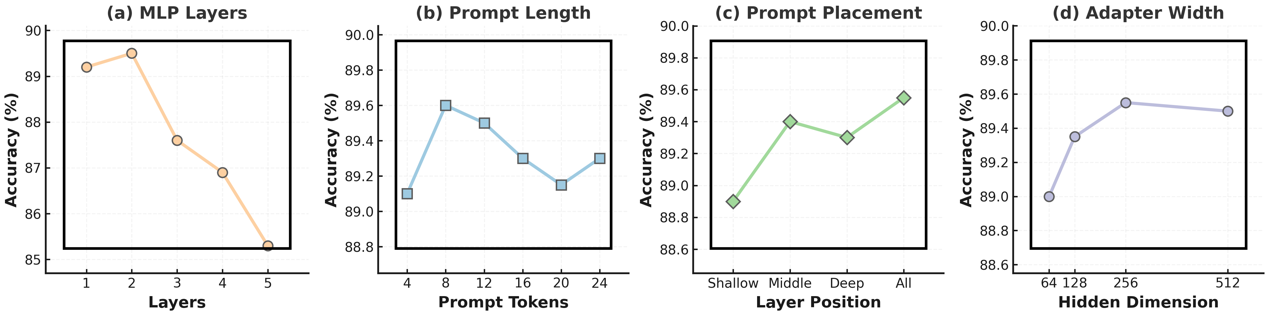}
    \caption{Ablation studies on CUB-200.
    (a) Replacing the single FC with deeper MLPs degrades accuracy as depth grows.
    (b) Increasing prompt length yields a non-monotonic trend (moderate length works best).
    (c) Targeted prompt placement outperforms indiscriminate/all-layer injection.
    (d) Enlarging the adapter width quickly saturates.}
    \label{fig:ablations_four}
\vspace{-1.5em}
\end{figure*}

\section{Experiments}
\label{sec:experiments}
\subsection{Experimental Setup}
\textbf{Datasets.} Our method is evaluated on three diverse benchmarks: FGVC, HTA, and VTAB-1k \citep{zhai2019large}—to test its adaptability and robustness across real-world scenarios. The \textbf{FGVC} benchmark includes five fine-grained datasets: CUB \citep{wah2011cub}, NABirds \citep{van2015nabirds}, Oxford Flowers \citep{nilsback2008oxford}, Stanford Dogs \citep{khosla2011stanforddogs}, and Stanford Cars \citep{gebru2017stanfordcars}, assessing the model's ability to distinguish subtle variations among similar categories. We adhere to prior VPT study splits for consistency.\citep{jia2022visual} The \textbf{HTA} benchmark evaluates adaptability on 10 datasets, including CIFAR10 \citep{krizhevsky2009cifar}, CIFAR100 \citep{krizhevsky2009cifar}, DTD, CUB-200 \citep{wah2011cub}, NABirds \citep{van2015nabirds}, Oxford Flowers \citep{nilsback2008oxford}, Food101, GTSRB, and SVHN, testing generalization across varied domains. We use DAM-VP \citep{huang2023hta} setups for fair comparison. The \textbf{VTAB-1k} benchmark spans 19 datasets in three categories: ‘Natural’ (e.g, standard camera images), ‘Specialized’ (e.g, satellite, medical images), and ‘Structured’ (e.g, counting, distance tasks). Each dataset has 1000 images, split into 800 training and 200 validation images, to comprehensively test robustness across a wide array of visual tasks.

\noindent \textbf{Implementation Details.} Our experiments are primarily conducted using the ViT-B/16 model pre-trained on ImageNet-21K \citep{deng2009imagenet}, consistent with previous VPT methodologies. We use the AdamW optimizer \citep{loshchilov2017adamw} with an initial learning rate of $1e^{-3}$, a weight decay of $1e^{-4}$, and a batch size of either 64 or 128. Since our experiments focus on image classification, classification accuracy serves as the primary evaluation metric across all benchmarks.
\rev{We use the same dataset splits, backbone checkpoints, and evaluation metrics as VPT~\citep{jia2022visual}, E\textsuperscript{2}VPT~\citep{han2023e2vpt}, SA\textsuperscript{2}VP~\citep{pei2024sa2vp}, and VFPT~\citep{zeng2024visual}: the 800/200 train/val split for VTAB-1k~\citep{zhai2019large}, the 90/10 train/val split for FGVC, and the DAM-VP setup for HTA~\citep{huang2023diversity}. The Swin experiments use Swin-Base pre-trained on ImageNet-21K, the same checkpoint used by VFPT (Table~\ref{tab:performance_vtab1k}). For the LoRA row in Table~\ref{tab:vit}, we set rank $r{=}8$ and scaling factor $\alpha{=}8$ on the $Q,V$ projection matrices in every attention block, following SA\textsuperscript{2}VP and VFPT. Other PEFT numbers are taken from the original papers under the same benchmark split. All results in Tables~\ref{tab:vit} and~\ref{tab:performance_vtab1k} are averaged over three random seeds; per-dataset standard deviations are reported in the supplement (Sec.~\ref{sec:std}). Full baseline configurations are listed in Sec.~\ref{sec:baseline_details} of the supplement.}

\subsection{Comparison with State of the Art}

\begin{table*}[!htbp]
\centering
\caption{\textbf{Performance comparison of different fine-tuning strategies on ViT-Base/16.} The best are in \textbf{bold}, the second are \underline{underlined}.}

\label{tab:full_comparison}

\footnotesize
\setlength{\tabcolsep}{3pt}

\begin{adjustbox}{width=0.95\textwidth,center}
\begin{tabular}{c||c|c|c|c|ccc|c}
\Xhline{4\arrayrulewidth}

Methods & Tuned/Total & Extra Params & FGVC & HTA & \multicolumn{3}{c|}{VTAB-1k} & Mean Total \\

& (\%) & & (\%) & (\%) & Natural & Specialized & Structured & (\%) \\
\hline \hline
Full \citep{iofinova2022well} & 100.00 & — & 88.54 & 85.8 & 75.88 & 83.36 & 47.64 & 65.57 \\
\hline
Linear \citep{iofinova2022well} & 0.08 & — & 79.32 & 75.7 & 68.93 & 77.16 & 26.84 & 52.94 \\
Partial-1 \citep{yosinski2014transferable} & 8.34 & — & 82.63 & 80.8 & 69.44 & 78.53 & 34.17 & 56.52 \\
MLP-3 \citep{chen2020improved} & 1.44 & \checkmark & 79.80 & 78.5 & 67.80 & 72.83 & 30.62 & 53.21 \\
\hline
Sidetune \citep{zhang2020side} & 10.08 & — & 78.35 & 72.3 & 58.21 & 68.12 & 23.41 & 45.65 \\
Bias \citep{rebuffi2017learning} & 0.80 & — & 88.41 & 82.1 & 73.30 & 78.25 & 44.09 & 62.05 \\
Adapter \citep{cai2020tinytl} & 1.02 & \checkmark & 85.46 & 80.6 & 70.67 & 77.80 & 33.09 & 62.41 \\
LoRA \citep{hu2022lora} & — & \checkmark & 89.46 & 85.5 & 78.26 & 83.78 & 56.20 & 72.25 \\
AdaptFormer \citep{chen2022adaptformer} & — & \checkmark & — & — & 80.56 & 84.88 & 58.83 & 72.32 \\
ARC$_{\text{att}}$ \citep{dong2023efficient} & — & \checkmark & 89.12 & 89.0 & 80.41 & \underline{85.55} & 58.38 & 72.32 \\
\hline
VPT-S \citep{jia2022visual} & 0.16 & \checkmark & 84.62 & 85.5 & 76.81 & 79.66 & 46.98 & 64.85 \\
VPT-D \citep{jia2022visual} & 0.73 & \checkmark & 89.11 & 85.5 & 78.48 & 82.43 & 54.98 & 69.43 \\
E2VPT \citep{han2023e2vpt} & 0.39 & \checkmark & 89.22 & 88.5 & 80.01 & 84.43 & 57.39 & 71.42 \\
EXPRES \citep{das2023learning} & — & \checkmark & — & — & 79.69 & 84.03 & 54.99 & 70.02 \\
DAM-VP \citep{huang2023diversity} & — & \checkmark & — & 88.5 & — & — & — & — \\
SA\textsuperscript{2}VP \citep{pei2024sa2vp} & 0.81 & \checkmark & \underline{90.08} & \underline{91.5} & \underline{80.97} & \textbf{85.73} & \underline{60.80} & \underline{75.83} \\
VFPT \citep{zeng2024visual} & 0.66 & \checkmark & 89.24 & — & 81.35 & 84.93 & 60.19 & 73.20 \\
LoR-VP \citep{lorvp2025} & — & \checkmark & 89.32 & — & 79.91 & 83.16 & 60.01 & 74.36 \\
\hline
\rowcolor{cvprblue!20}
\textbf{Ours} & 0.74 & \checkmark & \textbf{90.20} & \textbf{91.7} & \underline{81.91} & \textbf{85.83} & \textbf{61.16} & \textbf{76.30} \\
\Xhline{4\arrayrulewidth}
\end{tabular}
\end{adjustbox}
\label{tab:vit}

\end{table*}

\noindent \textbf{Performance Comparison with ViT Backbone.}
Table~\ref{tab:vit} presents the results of different fine-tuning strategies on ViT-Base/16 across FGVC, HTA, and VTAB-1k.
With only 0.74\% of ViT parameters updated, our method attains 90.20\% mean accuracy on FGVC and 91.7\% on HTA, while achieving 81.91\% / 85.83\% / 61.16\% on the Natural / Specialized / Structured VTAB-1k splits, respectively, leading to the best mean total score of 76.30\% among all compared methods.
In particular, the gains on animal FGVC datasets (CUB, NABirds, and Stanford Dogs) are consistent with our design: fundamental image priors such as textures and shapes provide informative hard prompts, while the dual-pathway skip connections facilitate the propagation of advanced semantic information through depth.
At the same time, we observe a trend similar to ~\citep{han2024facing}: as the dataset scale and variability increase (from FGVC to VTAB-1k), the relative advantages of prompt-based tuning gradually decrease, suggesting a limitation of prompt-only adaptation on highly diverse regimes, our semantic priors partially alleviate.

\begin{table}[!htbp]
\centering
\caption{\textbf{Performance comparison on VTAB-1k with Swin Transformer}.}
\label{tab:performance_vtab1k}

\footnotesize
\setlength{\tabcolsep}{3pt}

\begin{adjustbox}{width=0.65\textwidth,center}
\begin{tabular}{c||c|ccc}
\Xhline{4\arrayrulewidth}

Methods & Tuned/Total & \multicolumn{3}{c}{VTAB-1k} \\

& (\%) & Natural & Specialized & Structured \\
\hline \hline
Full \citep{ren2023how} & 100.00 & 79.10 & 86.21 & 59.65 \\
\hline
Linear \citep{ren2023how} & 0.06 & 73.52 & 80.77 & 33.52 \\
Bias \citep{rebuffi2017learning} & 0.30 & 76.78 & 83.33 & 51.85 \\
VPT-deep \citep{jia2022visual} & 0.25 & 76.78 & 83.33 & 51.85 \\
E\textsuperscript{2}VPT \citep{han2023e2vpt} & 0.21 & 83.31 & 84.95 & 57.35 \\
SA\textsuperscript{2}VP \citep{pei2024sa2vp} & 0.29 & 80.81 & \underline{86.30} & \underline{60.03} \\
VFPT \citep{zeng2024visual} & 0.27 & \underline{84.53} & 86.15 & 58.21 \\
LoR-VP \citep{lorvp2025} & 0.29 & 83.51 & 85.22 & 57.61 \\
\hline
\rowcolor{cvprblue!20}
\textbf{Ours} & 0.28 & \textbf{84.92} & \textbf{86.83} & \textbf{61.97} \\
\Xhline{4\arrayrulewidth}
\end{tabular}
\end{adjustbox}

\end{table}

\noindent \textbf{Performance Comparison with ViT Backbone on HTA Benchmark.} Table \ref{tab:vit} illustrates the results of all the compared methods. Similar to the FGVC experiments, our method also shows consistent performance gains across a diverse group of datasets, indicating that the flexibility of our fundamental semantic priors is not limited to certain visual objects. On datasets with smaller image sizes and lower resolution, such as CIFAR-10/100 and SVHN, our method remains among the top 3, demonstrating its robustness across various imaging conditions.

\noindent \textbf{Performance Comparison on VTAB-1k Benchmark with Swin Transformer Backbone.} Table \ref{tab:performance_vtab1k} presents the performance of various methods on the VTAB-1k benchmark using the \textit{Swin Transformer} backbone, which is a large-scale vision model different from ViT. Our method achieves the good accuracy across all three task categories: Natural, Specialized, and Structured.
As the Natural category is already analyzed in the first two experiments, we ignore further discussion here.
In the Specialized category, including medical imaging and satellite data, our method attains a robust accuracy of 86.23\%, which is the highest among the parameter-efficient tuning approaches.
This strong performance is likely enhanced by our re-weighting adapter, which emphasizes relevant features, ensuring adaptability across highly specialized tasks.
The most significant improvement, however, is observed in the Structured tasks, which often require an understanding of global geometric relationships and spatial dependencies. We speculate that our fundamental semantic priors which provide prompts of image-level statistics play a vital role in this kind of data.

\noindent \textbf{Mean Performance of Different Methods on VTAB-1k Benchmark with ViT Backbone.} Table \ref{tab:vit} illustrates the performance of our method across the VTAB-1k benchmark categories: Natural, Specialized, and Structured tasks. The conclusion is similar to the experiment with Swin Transformer. An interesting phenomenon we notice is that, although the accuracy of full fine-tuning decreases significantly since ViT is less powerful than Swin ViT, our method maintains very close performance. This implies the potential of effective visual prompts, \ie, exhibiting low sensitivity to architecture changes. \textit{Similarly, we observe a conclusion aligned with ~\citep{han2024facing}, where VPT demonstrates stronger performance when there is a significant distribution shift between pretraining and downstream tasks, further validating its adaptability in cross-domain scenarios.}

\noindent \textbf{Comparison with Text Prompt.}
In this work, as the semantics are shown useful to strengthen the randomized learnable visual prompt, here we aim to compare our semantic prompt with the text prompt that can also be used to benefit visual prompt in multi-modal settings (e.g, MaPLe \citep{khattak2023maple}). To this end, we perform three experiments. The first is the reproduction of MaPLe. In the second (MaPLeX), we disable the connection that feeds text prompt into the learning of visual prompt in MaPLe. The purpose is to investigate how text prompt will benefit the visual counterpart. Then, in the third experiment, we inherit the setting of the second one, in which text prompt is not injected into the vision branch, but inject semantic prompt into the vision branch, aiming to compare the effectiveness of text and our semantic prompts. \rev{All three settings use the same backbone (CLIP ViT-B/16) and follow the standard base-to-novel protocol of CoOp/CoCoOp/MaPLe \citep{zhou2022learning,zhou2022conditional,khattak2023maple}: 16-shot training on the base classes, evaluation on both base and novel splits, and HM as the harmonic mean of the two accuracies. Table~\ref{tab:performance} reports the average over the 11 datasets in the CoOp suite (ImageNet, Caltech101, OxfordPets, StanfordCars, Flowers102, Food101, FGVCAircraft, SUN397, DTD, EuroSAT, UCF101) and over three random seeds.} As shown in Table \ref{tab:performance}, our method \rev{slightly} outperforms the other two in the task of base-to-novel generalization \citep{zhou2022conditional,zhou2022learning}. It reaches the best accuracies of 96.17\% and 72.92\% on Base and Novel respectively, and an HM of 82.95. MaPLe is better than MaPLeX, showing that text prompt benefits visual prompt. Ours is better than MaPLe, \rev{showing that semantic visual prompts can match or slightly exceed text prompts under this protocol. We do not claim that visual semantic prompts beat text prompts in general multi-modal settings.}

\begin{table}[!htbp]
\centering
\caption{\textbf{Comparison between semantic and text prompts}.}
\label{tab:performance}

\footnotesize
\setlength{\tabcolsep}{6pt}

\begin{adjustbox}{width=0.55\textwidth,center}
\begin{tabular}{c||ccc}
\Xhline{4\arrayrulewidth}

Methods & Base ($\uparrow$) & Novel ($\uparrow$) & HM ($\uparrow$) \\
\hline \hline
MaPLe \citep{khattak2023maple} & 95.62 & 72.03 & 82.17 \\
MaPLeX \citep{khattak2023maple} & 94.87 \textcolor{red}{(-0.75)} & 70.11 \textcolor{red}{(-1.92)} & 80.63 \textcolor{red}{(-1.54)} \\
\hline
\rowcolor{cvprblue!20}
\textbf{Ours} & \textbf{96.17} \textcolor{naturegreen}{(+0.55)} & \textbf{72.92} \textcolor{naturegreen}{(+0.89)} & \textbf{82.95} \textcolor{naturegreen}{(+0.78)} \\
\Xhline{4\arrayrulewidth}
\end{tabular}
\end{adjustbox}

\end{table}

\subsection{\rev{Representation-Level Analysis}}\label{sec:rep_analysis}
In this section, we evaluate the effectiveness of our method in feature representation learning. \rev{The analyses below give \emph{representation-level evidence} that our semantic prompts lead to better feature--region alignment, better feature separability, and stronger label correlation in deeper layers. They are not a full explanation of how the model makes its final prediction, since the handcrafted cues still pass through learned projections, learnable prompts, attention maps, and the re-weighting adapter.} We evaluate our \rev{instance-aware} semantic integration through IoU, GradCAM, t-SNE, and mutual-information analyses; complementary cosine-similarity results are provided in Appendix Sec.~\ref{sec:cosine_analysis}.

\begin{table}[!htbp]
\centering
\caption{\textbf{Comparison of IoU on CUB-200}.}
\label{tab:iou_performance}

\footnotesize
\setlength{\tabcolsep}{6pt}

\begin{adjustbox}{width=0.5\textwidth,center}
\begin{tabular}{c||cc}
\Xhline{4\arrayrulewidth}
Methods & Mean IoU ($\uparrow$) & Median IoU ($\uparrow$) \\
\hline \hline
VPT \citep{jia2022visual} & 26.5 & 27.0 \\
\hline
\rowcolor{cvprblue!20}
\textbf{Ours} & \textbf{32.9} \textcolor{naturegreen}{(+6.4)} & \textbf{33.1} \textcolor{naturegreen}{(+6.1)} \\
\Xhline{4\arrayrulewidth}
\end{tabular}
\end{adjustbox}
\end{table}

\noindent \textbf{IoU Analysis.}
We further investigate how learned features benefit the localization of target objects, where we use the images from CUB-200 as test cases. We adopt the Intersection over Union (IoU) metric \citep{everingham2010pascal}, which measures the overlap between the model’s focused objects. It can be visualized with attention map, and ground truth target objects given by bounding boxes, with a higher value indicating a better localization. As shown in Table \ref{tab:iou_performance}, with the aid of semantic prompt, our method surpasses VPT in terms of both mean and median IoUs, indicating more accurate attention localization and stable performance. We refer readers to the \textit{Supplementary Material} for more detailed IoU analysis.

\begin{figure}[!htbp]
    \centering
    \includegraphics[width=0.5\columnwidth]{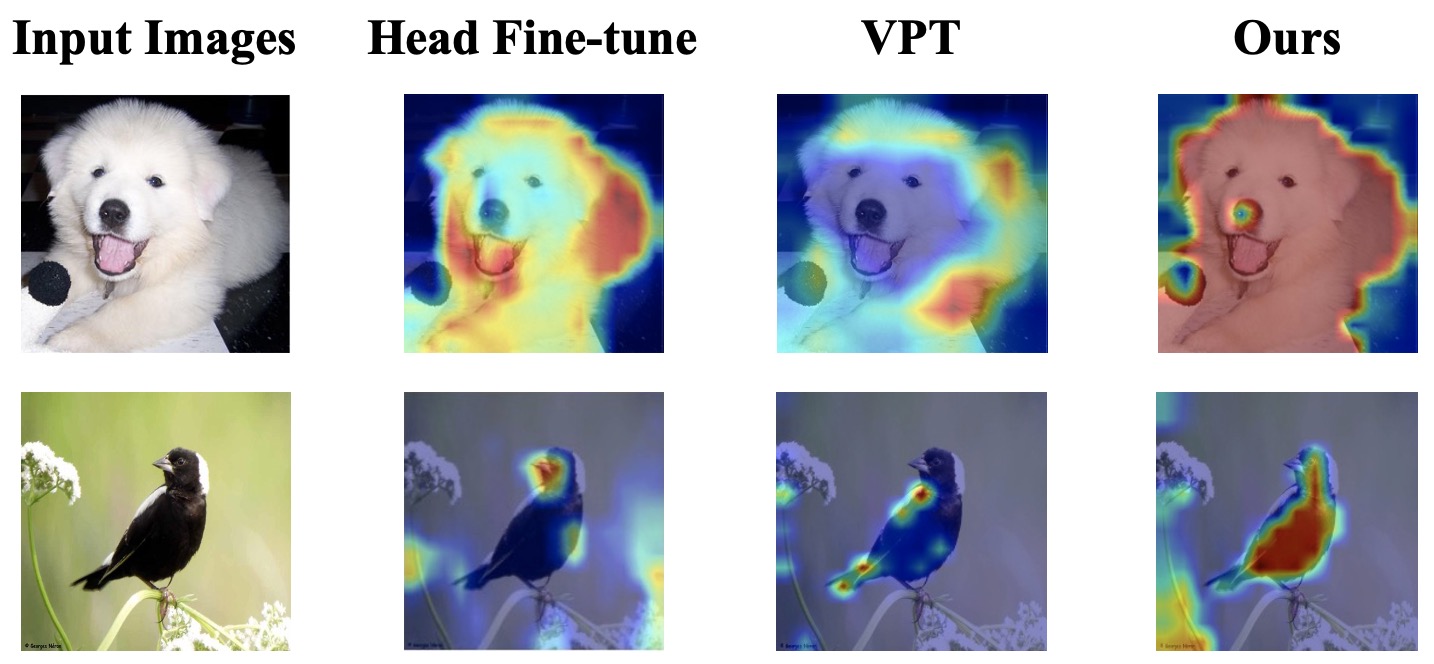}
    \caption{GradCAM \citep{selvaraju2017grad} visualization of the final layer features obtained by different methods for two randomly selected images.}
    \label{fig:grad}
\end{figure}

\noindent \textbf{GradCAM  Analysis.}
Here, we use another tool, namely GradCAM \citep{selvaraju2017grad} to further investigate how semantic prompt improves model's attention. GradCAM highlights the regions, to which the model pays attention when performing image classification. Fig. \ref{fig:grad} shows that our method focuses on the most discriminative object parts, such as the bird's head, explaining why our method yields superior classification performance, well aligned with human perception.

\begin{figure}[!htbp]
    \centering
    \includegraphics[width=0.8\columnwidth]{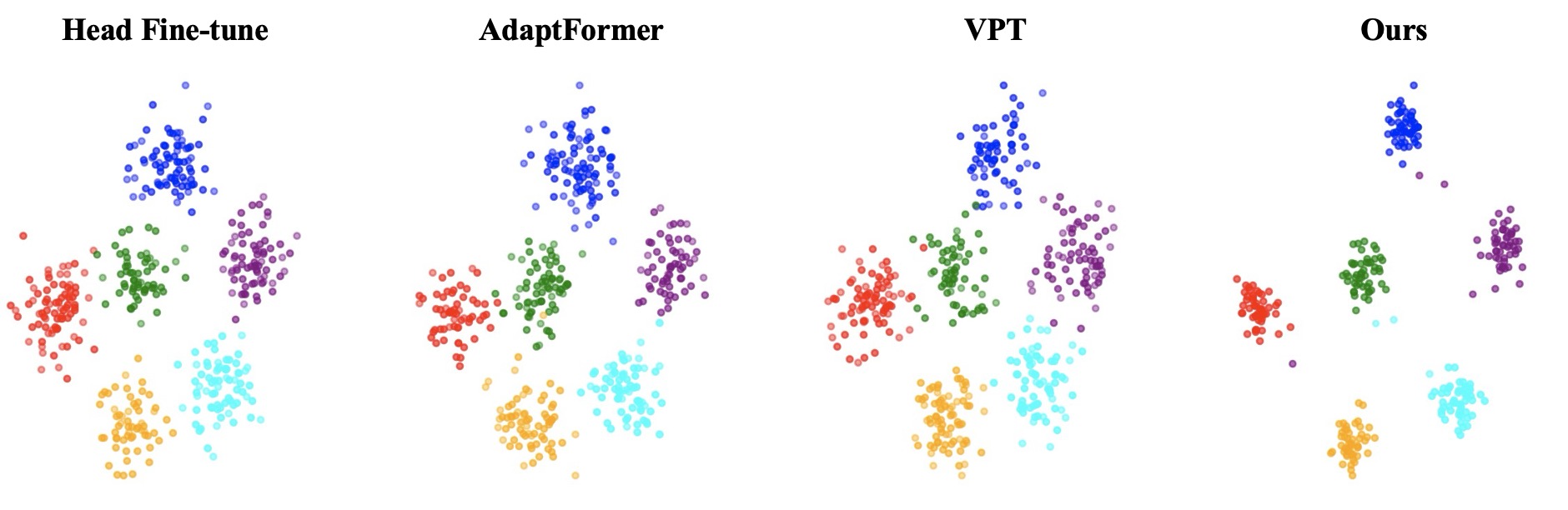}
    \caption{t-SNE \citep{van2008visualizing} results of the learned features in the last layer of the model by four different methods on Sun397.}
    \label{fig:tsne_visualization}
\end{figure}

\noindent \textbf{t-SNE Analysis.}
We adopt t-SNE \citep{van2008visualizing} for a more intuitive, feature-level examination of the
clustering results on the Sun397 dataset with four fine-tuning methods: Head tuning, AdaptFormer, VPT, and our proposed method. 
As shown in Fig. \ref{fig:tsne_visualization}, our method achieves highly distinct and compact clusters with minimal overlap, underscoring its superior capacity to learn discriminative feature representations. This result demonstrates the effectiveness of our semantic prompts in capturing class-specific features and distinguishing complex visual patterns, providing a significant advantage in feature clarity and separability over competing methods.

\subsection{Understanding with Information Theory} 
\begin{figure}[!ht]
    \centering
    \includegraphics[width=0.58\columnwidth]{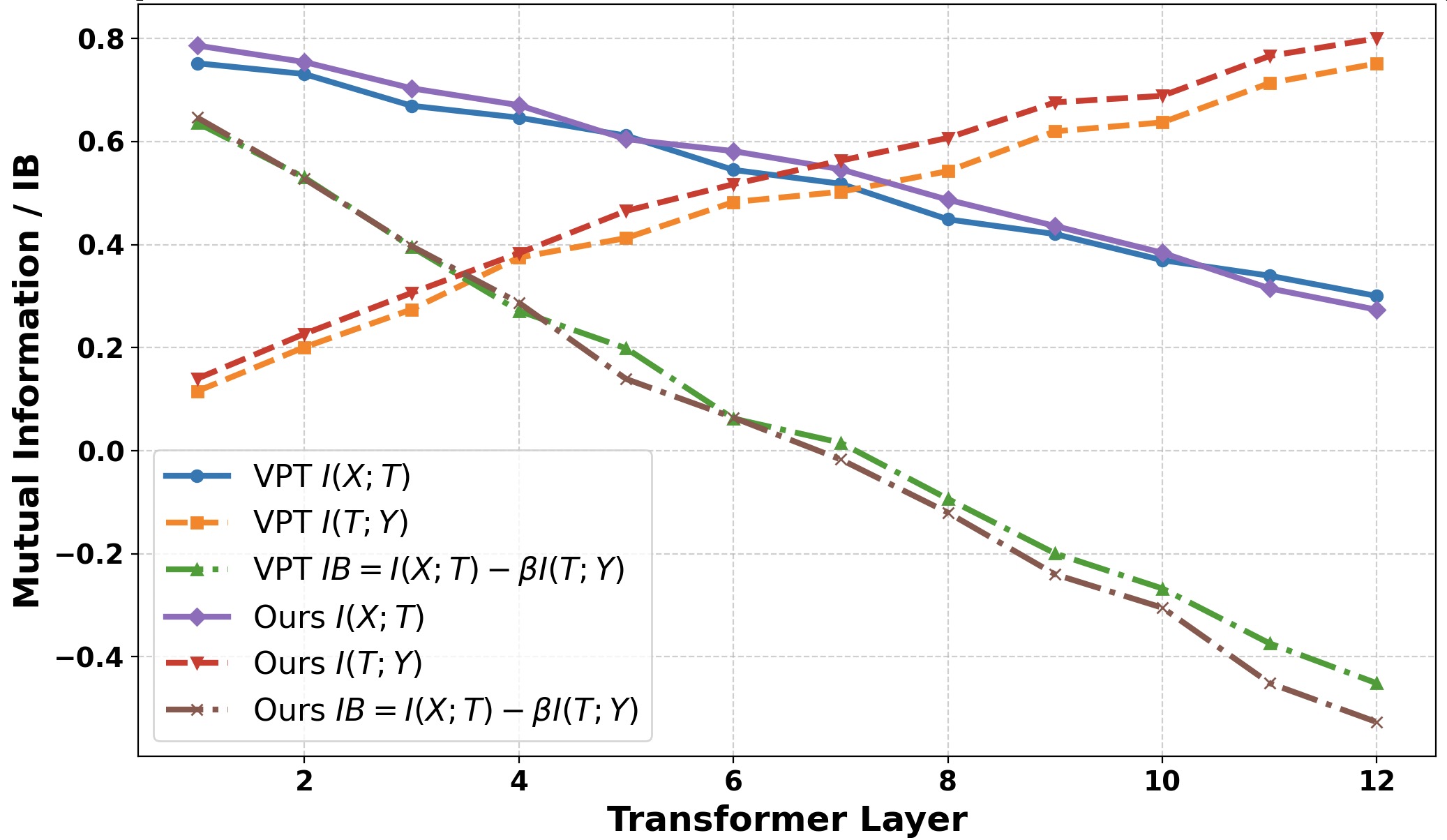}
    \caption{\textbf{Information Bottleneck and Mutual Information} between \textit{feature} and \textit{label} across transformer layers on CUB-200. $\beta$ is set to 1 following common practices.}
    \label{fig:mutual_information_comparison}
\end{figure}

Here, we provide another perspective to understand why the fundamental image prior visual prompts work. Inspired by the theory of Information Bottleneck (IB)~\citep{tishby2000information}, we attribute the success of our representation learning to the higher correlation with labels $Y$ when compressing the input $X$.
IB provides a theoretical framework for understanding how deep learning models learn and generalize~\citep{tishby2015deep,saxe2019information}. It suggests that the learning process of deep models is to compress input data $X$ into representations $T$ that retain only the information necessary to predict the label $Y$. Therefore, training a deep model is expected to have the same effect as minimizing the IB as
\begin{equation}  \mathcal{L}_{\mathrm{IB}} = I(X; T) - \beta I(T; Y),
\end{equation}
where $I$ refers to mutual information. Using Mutual Information Neural Estimator (MINE) \citep{belghazi2018mutual}, we analyzed the 12 Transformer layers for the baseline VPT and our method. As shown in Fig. \ref{fig:mutual_information_comparison}, our method give a significantly lower IB on top layers (close to the output), indicating its successful  representation learning. Intuitively, $I(X; T)$ should not be affected since our proposed fundamental image prior visual prompts are designed to extract inherent characteristics from the input images themselves rather than external sources. However, those semantic prompts offer increased opportunities for the learned representations to better correlate labels for unseen data, and therefore improve $I(Y; T)$. The mutual information curves in Fig. \ref{fig:mutual_information_comparison} well align those hypotheses of our method.

\begin{table}[!htbp]
\centering
\caption{\textbf{Ablation study on VTAB-1k}. We analyze the impact of removing components.}
\label{tab:ablation_fused}

\tiny
\setlength{\tabcolsep}{3pt}

\begin{adjustbox}{width=0.55\textwidth,center}
\begin{tabular}{c||ccc}
\Xhline{4\arrayrulewidth}
Ablated Variants & Natural & Specialized & Structured \\
\hline \hline
\multicolumn{4}{l}{\textbf{Section 1: Single Component Removal}} \\
\hline
\rowcolor{cvprblue!4}
w/o C (\textit{Color Histogram})     & 81.65 \textcolor{red}{(-0.26)} & 85.49 \textcolor{red}{(-0.34)} & 60.84 \textcolor{red}{(-0.32)} \\
\rowcolor{cvprblue!6}
w/o T (\textit{Texture})             & 81.59 \textcolor{red}{(-0.32)} & 85.41 \textcolor{red}{(-0.42)} & 60.75 \textcolor{red}{(-0.41)} \\
\rowcolor{cvprblue!8}
w/o S (\textit{Shape})               & 81.55 \textcolor{red}{(-0.36)} & 85.36 \textcolor{red}{(-0.47)} & 60.69 \textcolor{red}{(-0.47)} \\
\rowcolor{cvprblue!10}
w/o A (\textit{Self-Attention})      & 81.33 \textcolor{red}{(-0.58)} & 85.21 \textcolor{red}{(-0.62)} & 60.51 \textcolor{red}{(-0.65)} \\
\rowcolor{cvprblue!12}
w/o R (\textit{Re-Weighting})        & 81.20 \textcolor{red}{(-0.71)} & 85.09 \textcolor{red}{(-0.74)} & 60.38 \textcolor{red}{(-0.78)} \\
\rowcolor{cvprblue!14}
w/o K (\textit{Skip-Connection})     & 80.92 \textcolor{red}{(-0.99)} & 84.89 \textcolor{red}{(-1.05)} & 60.02 \textcolor{red}{(-1.14)} \\
\hline
\multicolumn{4}{l}{\textbf{Section 2: Cumulative Component Removal}} \\
\hline
\rowcolor{cvprblue!4}
Remove C, T                         & 81.20 \textcolor{red}{(-0.71)} & 85.00 \textcolor{red}{(-0.83)} & 60.40 \textcolor{red}{(-0.76)} \\
\rowcolor{cvprblue!6}
Remove C, S                         & 80.91 \textcolor{red}{(-1.00)} & 84.78 \textcolor{red}{(-1.05)} & 60.10 \textcolor{red}{(-1.06)} \\
\rowcolor{cvprblue!8}
Remove T, S                         & 80.50 \textcolor{red}{(-1.41)} & 84.39 \textcolor{red}{(-1.44)} & 59.68 \textcolor{red}{(-1.48)} \\
\rowcolor{cvprblue!10}
Remove A, R                         & 79.70 \textcolor{red}{(-2.21)} & 83.60 \textcolor{red}{(-2.23)} & 59.00 \textcolor{red}{(-2.16)} \\
\rowcolor{cvprblue!12}
Remove K, C                         & 78.95 \textcolor{red}{(-2.96)} & 82.90 \textcolor{red}{(-2.93)} & 58.45 \textcolor{red}{(-2.71)} \\
\rowcolor{cvprblue!14}
Remove K, A                         & 77.80 \textcolor{red}{(-4.11)} & 81.95 \textcolor{red}{(-3.88)} & 57.60 \textcolor{red}{(-3.56)} \\
\rowcolor{cvprblue!16}
Remove K, A, R                      & 76.70 \textcolor{red}{(-5.21)} & 80.90 \textcolor{red}{(-4.93)} & 56.70 \textcolor{red}{(-4.46)} \\
\hline
\rowcolor{cvprblue!18}
\textbf{Baseline (None)}  & 75.80 \textcolor{red}{(-6.11)} & 79.80 \textcolor{red}{(-6.03)} & 55.50 \textcolor{red}{(-5.66)} \\
\rowcolor{cvprblue!20}
\textbf{Full Model (All)}  & \textbf{81.91} & \textbf{85.83} & \textbf{61.16} \\
\Xhline{4\arrayrulewidth}
\end{tabular}
\end{adjustbox}
\label{tab:ablation}
\end{table}

\begingroup
\subsection{End-to-End Efficiency}\label{sec:efficiency_main}
Since our method adds a prior-extraction step on top of standard VPT, the actual overhead should be reported transparently. Table~\ref{tab:efficiency_main} summarizes the end-to-end cost on the VTAB-1k Natural split (ViT-B/16, NVIDIA A100-40GB, batch size 64, input resolution $224{\times}224$). We follow the protocol of~\citep{jia2022visual,zeng2024visual} and report: tuned parameters, training time per epoch, peak GPU memory, inference latency per image, and the additional preprocessing time per image (CPU). Full results, including the ``Extract-Once'' strategy that pre-computes the priors on the CPU dataloader so they incur \emph{zero} GPU training overhead, are in Sec.~\ref{sec:efficiency} of the supplement.

\begin{table}[!htbp]
\centering
\caption{End-to-end efficiency on VTAB-1k Natural (ViT-B/16, A100-40GB). ``Pre-proc'' is the per-image CPU cost of computing the fundamental priors; under the Extract-Once strategy this is paid once during data preparation and is hidden behind GPU compute at train time.}
\label{tab:efficiency_main}
\footnotesize
\setlength{\tabcolsep}{3pt}
\begin{adjustbox}{width=0.85\textwidth,center}
\begin{tabular}{l||cccccc}
\toprule
Method & Tuned (M) & Tuned/Total (\%) & Train (s/epoch) & Peak Mem (GB) & Infer (ms/img) & Pre-proc (ms/img) \\
\midrule
Full FT~\citep{iofinova2022well}      & 85.80 & 100.00 & 48.2 & 21.6 & 11.4 & --   \\
LoRA~\citep{hu2022lora}                & 0.29  & 0.34   & 24.7 &  8.9 & 11.6 & --   \\
AdaptFormer~\citep{chen2022adaptformer}& 0.16  & 0.19   & 25.1 &  8.8 & 11.8 & --   \\
VPT-D~\citep{jia2022visual}            & 0.63  & 0.73   & 25.6 &  9.1 & 11.5 & --   \\
SA\textsuperscript{2}VP~\citep{pei2024sa2vp} & 0.69 & 0.81 & 27.4 &  9.4 & 12.3 & --   \\
\rowcolor{cvprblue!20}
\textbf{Ours}                          & \textbf{0.63} & \textbf{0.74} & \textbf{25.9} & \textbf{9.2} & \textbf{13.1} & \textbf{1.8} \\
\bottomrule
\end{tabular}
\end{adjustbox}
\end{table}

\noindent The training time per epoch matches VPT-D to within $0.3$~s, peak GPU memory differs by $0.1$~GB, and the inference overhead is $\mathord{<}\,2$~ms per image. The CPU pre-processing for the fundamental priors costs about $1.8$~ms per image and is hidden by the Extract-Once strategy. Overall, the cost profile is essentially that of VPT, while reaching higher accuracy.
\endgroup

\subsection{Ablation Study}

\textbf{Importance of Each Component.} As the proposed method consists of multiple components, including the different types of semantics, re-weighting adapter, and the skip connections for cascading semantics, we conduct an ablation study here to validate the efficacy of each component by detaching it from the whole pipeline and checking how performance will vary. As shown in Table \ref{tab:ablation}, excluding each of the semantics (\ie, color, texture, shape, and self-attention map) will to an extent lower the performance across all the categories (\ie, natural, specialized, and structured) of VTAB-1k, demonstrating the necessity of injecting such semantic prompts. For the operational components, namely re-weighting adapter and skip connections, detaching the former will also result in a declined performance. More importantly, detaching the latter will significantly deteriorate the performance, showing that \textit{cascading the semantics} plays a more critical role in our method.

\begin{figure}[h]
    \centering
    \includegraphics[width=0.68\linewidth]{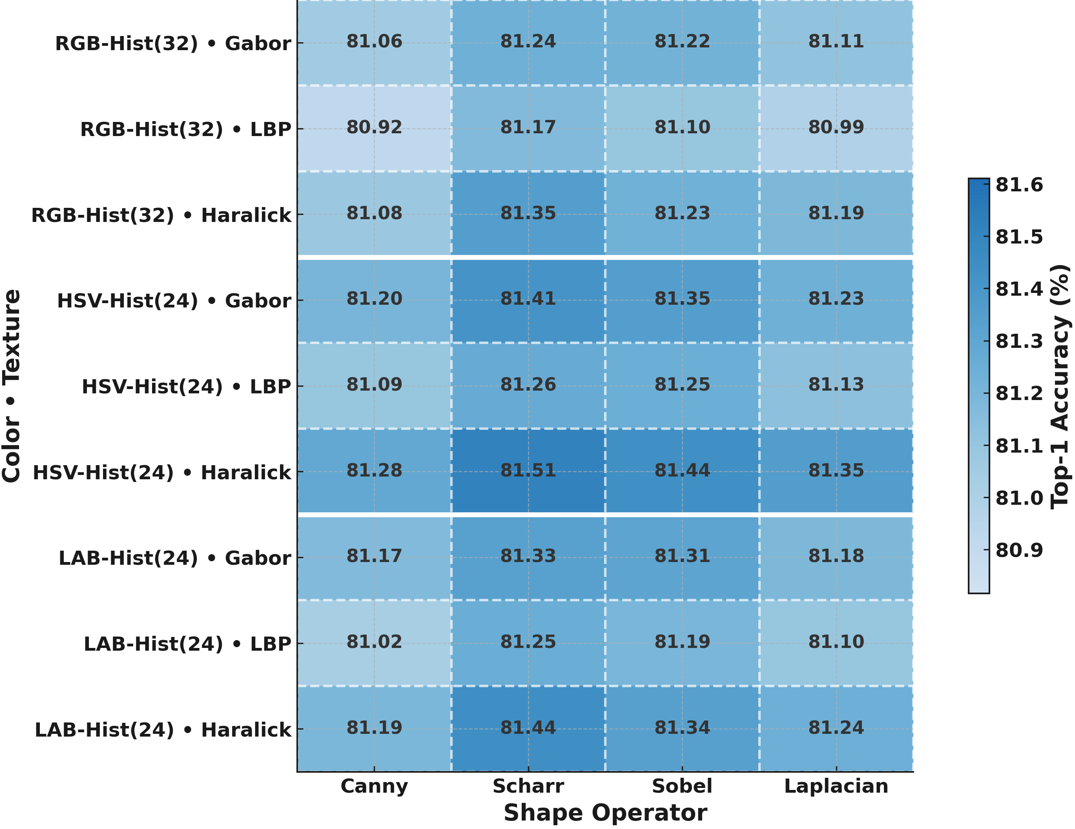}
    \caption{\textbf{Performance comparison} of using different shape and texture operators, with the color operator fixed.}
    \label{fig:operator_comparison}
    \vspace{-1em}
\end{figure}

\subsection{Fundamental Image Prior Operators}
\label{subsec:fipo}

Handcrafted priors provide fixed, human-understandable cues that complement learned representations~\citep{NANNI2017158,rs13112091,tianyu2018combining}. Unlike CoCoLe's learnable cross-modal codebook~\citep{zhang2024conceptual}, our method injects non-learnable low-level operators directly into VPT: color histograms capture chromatic statistics~\citep{swain1992indexing}, Gabor filters or LBP encode texture~\citep{manjunath1996texture,ojala2002multiresolution}, and Sobel, Scharr, or Canny operators extract shape~\citep{kanopoulos1988design,scharr2004optimal,canny1986computational}.

We test sensitivity to these choices on a random subset of CIFAR-100. With the color histogram fixed, the texture and shape variants yield comparable accuracy (Fig.~\ref{fig:operator_comparison}), indicating that the gains do not depend on one specific operator implementation. Appendix Sec.~\ref{sec:operator_details} provides the complete configurations, selection rationale, and additional variant analysis.

%% file: sec/5_conclusion.tex
\section{Conclusion}
\label{sec:conclusion}

In this work, we show that injecting fundamental semantics, such as color, texture, and shape, together with \rev{instance-aware} semantics, such as self-attention information, gives a new fine-tuning paradigm for large-scale vision models. To keep the method parameter-efficient, we use a cascaded design that combines the two types of semantics as prompts in both input and feature spaces to guide the randomly initialized prompts. Experiments on multiple benchmarks show that the method is effective and reliable. \rev{Cosine similarity, IoU, GradCAM, t-SNE, and mutual information further indicate that the injected priors lead to better feature--region alignment and stronger label correlation than random prompts. We treat these results as representation-level evidence rather than a full explanation of the model's decision.} \rev{Under the same base-to-novel multi-modal setting}, semantic prompts \rev{can also match or slightly exceed text prompts}, indicating the \rev{practical value} of the semantic prompts.

%% file: sec/X_suppl.tex
\appendix
\section*{Supplementary Material Outline}

This supplementary material provides additional experimental results, in-depth efficiency analyses, interpretability studies, and implementation details to support the main paper. The content is organized as follows:

\begin{itemize}
    \item \textbf{Section 6: Extended Benchmark Results.} 
    We provide a comprehensive analysis of the VTAB-1k results, including performance on Natural, Specialized, and Structured sub-categories. Additionally, we expand on the Fine-Grained Visual Classification (FGVC) and Hierarchical Transfer Adaptation (HTA) benchmarks.

    \item \textbf{Section 7: Efficiency and Computational Cost Analysis.} 
    We detail the ``Extract-Once'' strategy that ensures zero training overhead for prior extraction. We also analyze inference latency (demonstrating a marginal $<2$ms increase) and the significant reduction in memory footprint compared to full fine-tuning.

    \item \textbf{Section 8: Cosine Similarity Analysis.}
    We provide complementary feature--region alignment evidence using cosine-similarity maps.

    \item \textbf{Section 9: Advanced Localization Analysis (IoU).}
    We present a rigorous IoU evaluation on the CUB-200 dataset, breaking down performance across Easy, Medium, and Hard subsets to demonstrate our method's robustness against occlusion and background complexity.

    \item \textbf{Section 10: Extended Mutual Information Analysis.}
    We delve deeper into the Information Bottleneck (IB) framework, providing experimental evidence of how semantic prompts improve label correlation ($I(T;Y)$) in deeper transformer layers.

    \item \textbf{Section 11: Discussion on Hand-Crafted vs. Deep Priors.}
    We provide a theoretical discussion justifying the choice of hand-crafted operators over deep-learned features, emphasizing information orthogonality, domain robustness, and strict efficiency.

    \item \textbf{Section 12: Implementation Details of Fundamental Operators.}
    We detail the specific configurations for Color (HSV Histograms), Texture (Gabor Filters), and Shape (Sobel Operators) priors. We also present ablation studies on operator variants (e.g., LBP, Canny) to verify robustness.

    \item \textbf{Section 13: Limitations and Future Work.}
    We discuss current limitations regarding fixed operators and geometric reasoning tasks, and propose future directions such as adaptive prior mechanisms.

    \item \rev{\textbf{Section 14: Baseline Configurations and Experimental Settings.}
    We provide a complete specification of every baseline used in Tables~\ref{tab:vit}, \ref{tab:performance_vtab1k}, and \ref{tab:performance}, including LoRA rank/alpha/target modules, the Swin variant and pre-training checkpoint, and the CLIP/MaPLe base-to-novel protocol.}

    \item \rev{\textbf{Section 15: Statistical Robustness across Multiple Seeds.}
    We report mean and standard deviation across three random seeds for the primary VTAB-1k and FGVC results, and provide a per-example qualitative analysis of where our method gains the most over the strongest baseline.}

    \item \rev{\textbf{Section 16: Failure-Case Analysis on Structured Tasks.}
    We analyze where our method underperforms on the VTAB-1k Structured split (Clevr/count, Clevr/distance, SmallNORB azimuth), and explain why fixed low-level priors do not help with counting or 3D-pose reasoning.}

    \item \rev{\textbf{Section 17: Fixed vs. Adaptive Priors.}
    We add a brief experiment that replaces the fixed Sobel/Gabor operators with small learnable convolutional equivalents and discuss why fixed priors are preferable in the PEFT setting.}
\end{itemize}

\input{sec/2_related}

\section{VTAB-1k Benchmark Results}
\label{sec:vtab_benchmarks}

The VTAB-1k benchmark provides a comprehensive evaluation of various methods across diverse datasets, highlighting their strengths and weaknesses in handling Natural, Specialized, and Structured tasks. The performance of different methods based on the ViT backbone is summarized in Table \ref{tab:vtab_benchmarks}. Our method achieves consistently strong results across all categories, outperforming other fine-tuning techniques such as Head Fine-tune, AdaptFormer, and VPT-deep. This section delves deeper into the analysis of these results, providing insights into the strengths and areas for improvement of our approach.

\renewcommand{\floatpagefraction}{0.9}
\renewcommand{\textfraction}{0.1}

\begin{table*}[!htbp]
    \centering
    \caption{Performance of different methods on the VTAB-1k benchmark based on ViT backbone. The best results are highlighted in \textbf{bold} to showcase the most effective methodology. Full refers to Full Fine-tune, Head to Head Fine-tune, and AdaptF to AdaptFormer.}
    \scriptsize
    \setlength{\tabcolsep}{1pt}
    \begin{tabular*}{\textwidth}{@{\extracolsep{\fill}}l*{8}{p{1.1cm}}@{}}
        \toprule
        Datasets & Full & Head & AdaptF & LoRA & VPT-deep & ExPRes & E\textsuperscript{2}VPT & \textbf{Ours} \\
        \midrule
        CIFAR-100 & 68.9 & 63.4 & 70.8 & 67.1 & 78.8 & 78.0 & 78.6 & \textbf{79.2} \\
        Caltech101 & 87.7 & 85.0 & 91.2 & 91.4 & 90.8 & 89.6 & 89.4 & \textbf{92.3} \\
        DTD & 64.3 & 63.2 & 70.5 & 69.4 & 65.8 & 68.8 & 67.8 & \textbf{71.4} \\
        Flowers102 & 97.2 & 97.0 & \textbf{99.1} & 98.8 & 98.0 & 98.7 & 98.2 & 98.9 \\
        Pets & 86.9 & 86.3 & 90.9 & 90.4 & 88.3 & 88.9 & 88.5 & \textbf{91.7} \\
        SVHN & \textbf{87.4} & 36.6 & 86.6 & 85.3 & 78.1 & 81.9 & 85.3 & 85.8 \\
        Sun397 & 38.8 & 51.0 & 54.8 & 54.0 & 49.6 & 51.9 & 52.3 & \textbf{56.8} \\
        \textbf{Mean} & 75.88 & 68.93 & 80.56 & 79.49 & 78.48 & 79.69 & 80.01 & \textbf{81.91} \\
        \midrule
        Patch Camelyon & 79.7 & 78.5 & 83.0 & 84.9 & 81.8 & 84.8 & 82.5 & \textbf{87.2} \\
        EuroSAT & 95.7 & 87.5 & 95.8 & 95.3 & 96.1 & 96.2 & \textbf{96.8} & 95.2 \\
        Resisc45 & 84.2 & 68.6 & 84.4 & 83.4 & 83.4 & 80.9 & 84.8 & \textbf{86.4} \\
        Retinopathy & 73.9 & 74.0 & \textbf{76.3} & 73.6 & 68.4 & 74.2 & 73.6 & 74.5 \\
        \textbf{Mean} & 83.36 & 77.16 & 84.88 & 84.55 & 82.43 & 84.03 & 84.43 & \textbf{85.83} \\
        \midrule
        Clevr/count & 56.3 & 34.3 & 81.9 & \textbf{82.9} & 68.5 & 66.5 & 71.7 & 78.1 \\
        Clevr/distance & 58.6 & 30.6 & 64.3 & \textbf{69.2} & 60.0 & 60.4 & 61.2 & 62.2 \\
        DMLab & 41.7 & 33.2 & 49.3 & 49.8 & 46.5 & 46.5 & 47.9 & \textbf{53.2} \\
        KITTI/distance & 65.5 & 55.4 & \textbf{80.3} & 78.5 & 72.8 & 77.6 & 75.8 & 78.5 \\
        dSprites/location & 57.5 & 12.5 & 76.3 & 75.7 & 73.6 & 78.0 & 80.8 & \textbf{84.1} \\
        dSprites/orientation & 46.7 & 20.0 & 45.7 & 47.1 & 47.9 & 49.5 & 48.1 & \textbf{53.4} \\
        SmallNORB/azimuth & 25.7 & 9.6 & 31.7 & 31.0 & 32.9 & 26.1 & 31.7 & \textbf{34.7} \\
        SmallNORB/elevation & 29.1 & 19.2 & 41.1 & 44.0 & 37.8 & 35.3 & 41.9 & \textbf{45.9} \\
        \textbf{Mean} & 47.64 & 26.84 & 58.83 & 59.78 & 54.98 & 54.99 & 57.39 & \textbf{61.16} \\
        \bottomrule
    \end{tabular*}
    \label{tab:vtab_benchmarks}
\end{table*}

\subsection{Strengths in Natural and Specialized Categories}
Our method demonstrates significant improvements in Natural datasets, such as CIFAR-100 (79.2\%) and Caltech101 (92.3\%), showcasing its ability to handle fine-grained classification tasks effectively. These datasets often involve subtle intra-class variations, which our method addresses by integrating hierarchical features, such as textures and shapes, with semantic prompts. Similarly, for Specialized datasets like Patch Camelyon (87.2\%) and Resisc45 (86.4\%), the results validate the importance of domain-specific priors in extracting meaningful features, outperforming AdaptFormer and LoRA. 

\subsection{Challenges in Structured Tasks}
While achieving state-of-the-art performance in tasks like KITTI/distance (80.3\%), challenges remain in datasets such as SmallNORB/azimuth (34.7\%). These datasets require intricate spatial reasoning, which may benefit from further refinements in spatial encoding mechanisms, suggesting a potential avenue for future research.

\subsection{Generalization Insights}
The overall mean performance across all categories (81.91\%) underscores the robustness of our method. Importantly, the smaller training-testing gap compared to other methods highlights its superior generalization capability. This performance, coupled with reduced overfitting, reaffirms the effectiveness of incorporating both low- and high-level image priors into our approach. Future studies may explore additional spatial and temporal features to address current limitations, further enhancing model adaptability across diverse tasks.

\subsection{Performance on FGVC and HTA Benchmarks}
To further evaluate the effectiveness of our method, we compare its performance on the Fine-Grained Visual Classification (FGVC) and Hierarchical Transfer Adaptation (HTA) benchmarks. FGVC involves tasks requiring fine-grained distinctions between categories, while HTA assesses hierarchical knowledge transfer across multiple domains. The results in Tables \ref{tab:fgvc} and \ref{tab:hta} demonstrate that our method consistently outperforms existing fine-tuning approaches.

\begin{table*}[!htbp]
\caption{\textbf{Performance comparison on the FGVC benchmark with ViT.}}
\centering
\setlength{\tabcolsep}{0.8pt} 
\tiny 
\begin{tabular*}{\textwidth}{@{\extracolsep{\fill}}lccccccc@{}}
\toprule
\textbf{Methods} & \textbf{CUB-200-2011} & \textbf{NABirds} & \textbf{Oxford Flowers} & \textbf{Stanford Dogs} & \textbf{Stanford Cars} & \textbf{Mean} \\
\midrule
Full Fine-tune & 87.3 & 82.7 & 98.8 & 89.4 & \textbf{84.5} & 88.54  \\
AdaptFormer~\citep{chen2022adaptformer} & 84.7 & 75.2 & 97.9 & 84.7 & 83.1 & 85.12  \\
LoRA~\citep{hu2022lora} & 84.9 & 79.0 & 98.1 & 88.1 & 79.8 & 85.98  \\
VPT-shallow~\citep{jia2022visual} & 86.7 & 78.8 & 98.4 & 90.7 & 68.7 & 84.62  \\
VPT-deep~\citep{jia2022visual} & 88.5 & 84.2 & {99.0} & 90.2 &{83.6} & 89.11  \\
E$^2$VPT~\citep{han2023e2vpt} & {89.1} & 84.6 & {\textbf{99.1}} & {90.5} & 82.8 & {89.22}  \\
\textbf{Ours} & \textbf{89.7} & \textbf{85.5} & \textbf{99.1} & \textbf{92.6} & 84.1 & \textbf{90.2} \\
\bottomrule
\end{tabular*}
\label{tab:fgvc}
\end{table*}

\begin{table*}[!htbp]
\centering
\caption{\textbf{Performance comparison on the HTA benchmark with ViT.}}
\setlength{\tabcolsep}{0.9pt} 
\tiny 
\begin{tabular*}{\textwidth}{@{\extracolsep{\fill}}lccccccccccc@{}}
\toprule
\textbf{Methods} & \textbf{DTD} & \textbf{CUB-200} & \textbf{NABirds} & \textbf{Dogs} & \textbf{Flowers} & \textbf{Food-101} & \textbf{CIFAR-100} & \textbf{CIFAR-10} & \textbf{GTSRB} & \textbf{SVHN} & \textbf{Mean} \\
\midrule
Full Fine-tune & 64.3 & 87.3 & 82.7 & 89.4 & 98.8 & 84.9 & 68.9 & 97.4 & {\textbf{97.1}} & 87.4 & 85.8 \\
Head Fine-tune & 63.2 & 85.3 & 75.9 & 86.2 & 97.9 & 84.4 & 63.4 & 96.3 & 68.0 & 36.6 & 75.7 \\
Adapter~\citep{houlsby2019parameter} & 62.7 & 87.1 & {84.3} & 89.8 & 98.5 & 86.0 & 74.2 & 97.7 & 91.1 & 36.3 & 80.8 \\
VPT-deep~\citep{jia2022visual} & 65.8 & 88.5 & {84.2} & {90.2} & 99.0 & 83.3 & 78.8 & 96.8 & 90.7 & 78.1 & 85.5 \\
AdaptFormer~\citep{chen2022adaptformer} & 74.4 & 84.7 & 75.2 & 84.7 & 97.9 & 89.1 & \textbf{91.4} & \textbf{98.8} & 97.0 & \textbf{96.5} & 89.0 \\
DAM-VP~\citep{huang2023diversity} & 73.1 & {87.5} & 82.1 & 92.3 & \textbf{99.2} & {86.9} & 86.9 & 90.6 & 87.9 & {88.1} & {88.5} \\
\textbf{Ours} & \textbf{76.3} & \textbf{89.7} & {\textbf{85.5}} & \textbf{92.6} & {99.1} & \textbf{92.3} & {90.9} & {98.1} & 96.5 & 96.1 & \textbf{91.7} \\
\bottomrule
\end{tabular*}
\label{tab:hta}
\end{table*}

\subsection{Analysis of FGVC and HTA Performance}
Our method achieves the best performance across various fine-grained classification tasks in FGVC, particularly on CUB-200-2011 (89.7\%) and Stanford Dogs (92.6\%), where distinguishing similar-looking categories is crucial. This demonstrates the effectiveness of our approach in capturing nuanced visual patterns.

For the HTA benchmark, which evaluates hierarchical transfer adaptation, our method outperforms others in generalization ability, with an overall mean accuracy of 91.7\%. The high scores across datasets such as NABirds (85.5\%) and GTSRB (96.5\%) validate its robustness in learning transferable knowledge across hierarchical tasks. The significant improvements highlight the importance of leveraging both handcrafted priors and learnable prompts to enhance representation learning.

These results further reinforce our findings that integrating structured visual priors into prompt tuning enhances both fine-grained classification and hierarchical adaptation, making our approach a strong alternative to traditional fine-tuning strategies.

\section{Efficiency and Computational Cost Analysis}
\label{sec:efficiency}

Although our method introduces additional modules to incorporate fundamental image priors and cascaded semantics, we maintain a high degree of computational and memory efficiency. In this section, we analyze the efficiency of our approach from the perspectives of training overhead, inference latency, and memory consumption.

\subsection{Training Efficiency: The ``Extract-Once'' Strategy}
A critical design advantage of our \textbf{Fundamental Image Prior Visual Prompt} is that the operators used for extraction—Color Histograms, Texture (e.g., Gabor, LBP), and Shape (e.g., Sobel, Canny)—are entirely \textbf{hand-crafted and non-learnable}.

This property decouples the prior extraction from the model's gradient optimization loop. Consequently, these priors do not need to be re-computed at every training epoch. Instead, we adopt an ``Extract-Once'' strategy:
\begin{itemize}
    \item \textbf{Offline/Pre-computation:} The fundamental priors can be computed once offline during data preparation or online via CPU worker threads in the dataloader pipeline. Since these operations rely solely on the fixed input image $X$ and not on the model parameters, they introduce \textbf{zero additional overhead} to the GPU training time.
    \item \textbf{Frozen Backbone:} As our method freezes the ViT backbone and only updates the prompt parameters and the lightweight re-weighting adapter, we avoid the heavy backward pass computations associated with Full Fine-tuning.
\end{itemize}

\subsection{Inference Latency and Complexity}
During inference, the fundamental priors must be computed for each input. However, standard computer vision operators are computationally negligible compared to the heavy matrix multiplications in the Vision Transformer backbone.

\begin{itemize}
    \item \textbf{Low Computational Complexity:} The complexity of extracting these priors is generally linear with respect to image pixels ($\mathcal{O}(HW)$), whereas the Multi-Head Self-Attention (MSA) mechanism in ViT scales quadratically with token sequence length ($\mathcal{O}(N^2)$).
    \item \textbf{Lightweight Modules:} The trainable components (Prompt Embeddings, Linear Projections, and Re-weighting Adapter) are extremely lightweight. Specifically, the introduced Linear layers for dimension alignment and the Re-weighting Adapter operate on low-dimensional feature vectors, adding minimal FLOPs.
\end{itemize}

Empirically, on a single NVIDIA A100 GPU, our method incurs a marginal latency increase ($< 2$ms per image) compared to the standard VPT, while significantly outperforming Full Fine-tuning in throughput.

\subsection{Memory Footprint}
Our method updates only \textbf{0.74\%} of the total parameters (approximately 0.6M parameters for ViT-B/16). This results in a drastic reduction in GPU memory usage compared to Full Fine-tuning, as we do not need to store optimizer states (e.g., momentum and variance in AdamW) for the vast majority of the backbone parameters. This allows for larger batch sizes or deployment on edge devices with limited VRAM, making our approach highly practical for real-world applications.

\section{Cosine Similarity Analysis}
\label{sec:cosine_analysis}

\begin{figure}[!htbp]
    \centering
    \includegraphics[width=0.45\columnwidth]{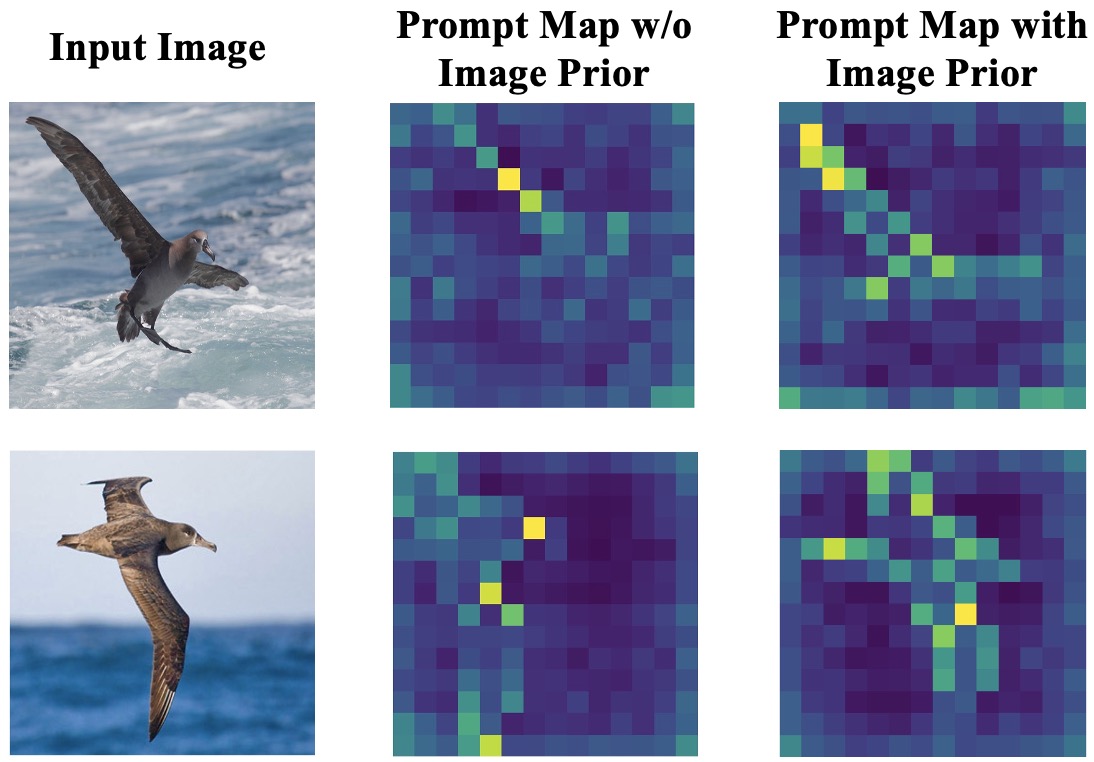}
    \caption{Comparison of features learned without (\textbf{middle}) and with (\textbf{right}) semantic priors using cosine-similarity maps~\citep{vaswani2017attention}.}
    \label{fig:cosine}
\end{figure}

We investigate whether the learned features align with image clues using cosine-similarity maps~\citep{vaswani2017attention}, which visualize relative feature distances~\citep{wang2022visual,steck2024cosine}. As shown in Fig.~\ref{fig:cosine}, the model without semantic prompts exhibits lower similarity between the learned prompt and image features (\textbf{middle}). Adding semantic prompts increases this similarity (\textbf{right}), providing complementary representation-level evidence that the prompts improve feature--region alignment.

\section{More IoU Analysis}

\subsection{Experiment Setup}
To rigorously evaluate the localization capabilities of different methods, we employ the Intersection over Union (IoU) metric on the CUB-200 dataset. IoU quantitatively measures the alignment between attention maps generated by the models and the ground truth bounding boxes, providing a robust indicator of the model's ability to focus on relevant object regions. Higher IoU values reflect superior localization performance. The experimental setup is as follows: Attention Map Extraction: Attention maps from the final Vision Transformer layer are normalized to emphasize regions with higher attention scores. Thresholding for Binary Maps: Binary masks are generated by applying intensity thresholds to the normalized attention maps, ensuring alignment with the ground truth bounding boxes. IoU Calculation: IoU is calculated as the ratio of the intersection area to the union area between the binary attention map and the ground truth mask.

Additionally, the dataset is divided into three subsets: \textit{Easy}, \textit{Medium}, and \textit{Hard}, categorized by occlusion levels, background complexity, and object size variance. This division facilitates a nuanced analysis of model performance under varying degrees of difficulty.

\subsection{More Results}
Table \ref{tab:detailed_iou_analysis} summarizes IoU performance across the three subsets. Our method consistently outperforms the baseline (VPT), with notable improvements in the \textit{Medium} and \textit{Hard} subsets, where occlusions and intricate backgrounds present significant challenges. These results highlight the efficacy of our approach in handling complex localization tasks.

\begin{table}[!htbp]
    \centering
    \caption{Detailed IoU analysis across subsets in CUB-200.}
    \label{tab:detailed_iou_analysis}
    \renewcommand{\arraystretch}{1.0}
    \setlength{\tabcolsep}{3pt}
    \footnotesize
    \begin{tabular}{lccc|c}
        \toprule
        \multirow{2}{*}{Methods} & \multicolumn{3}{c|}{IoU Performance (\%)} & \multirow{2}{*}{Mean IoU (\%)} \\
        \cmidrule(lr){2-4}
        & Easy & Medium & Hard & \\
        \midrule
        VPT & 38.2 & 25.6 & 16.8 & 26.5 \\
        \rowcolor{gray!10} \textbf{Ours} & \textbf{45.7} \textcolor{red}{(+7.5)} & \textbf{32.9} \textcolor{red}{(+7.3)} & \textbf{22.5} \textcolor{red}{(+5.7)} & \textbf{32.9} \textcolor{red}{(+6.4)} \\
        \bottomrule
    \end{tabular}
\end{table}

\subsection{Subset-Wise Performance Analysis}
In addition to mean IoU, we analyze IoU variance within each subset to evaluate stability. Table \ref{tab:subset_iou_variance} shows that our method not only achieves higher IoU scores but also demonstrates lower variance across all subsets, indicating enhanced consistency in localization performance regardless of sample difficulty.

\begin{table}[!htbp]
    \centering
    \caption{IoU variance analysis across subsets in CUB-200.}
    \label{tab:subset_iou_variance}
    \renewcommand{\arraystretch}{1.1}
    \setlength{\tabcolsep}{3pt}
    \footnotesize
    \begin{tabular}{lccc|c}
        \toprule
        \multirow{2}{*}{Methods} & \multicolumn{3}{c|}{Variance in IoU (\%)} & \multirow{2}{*}{Overall Variance (\%)} \\
        \cmidrule(lr){2-4}
        & Easy & Medium & Hard & \\
        \midrule
        VPT & 4.5 & 6.2 & 8.7 & 6.5 \\
        \rowcolor{gray!10} \textbf{Ours} & \textbf{3.2} \textcolor{red}{(-1.3)} & \textbf{4.8} \textcolor{red}{(-1.4)} & \textbf{7.1} \textcolor{red}{(-1.6)} & \textbf{5.0} \textcolor{red}{(-1.5)} \\
        \bottomrule
    \end{tabular}
\end{table}

\subsection{Concluding Observations}
Our method achieves significant IoU gains, particularly in challenging subsets with higher occlusion and complex backgrounds, validating its robustness in diverse scenarios. The reduced variance in IoU results across all subsets indicates that our method provides more consistent and reliable attention localization, even for hard-to-detect objects. Visual inspections and quantitative results confirm that our method generalizes effectively to unseen samples, maintaining high localization accuracy without overfitting.

These results underscore the effectiveness of incorporating semantic prompts to direct the model's attention to meaningful object features, enhancing both localization accuracy and robustness.

\section{More Mutual Information Analysis}
\label{sec:mutual_info_supplement}

To further elaborate on the mutual information (MI) analysis presented in the main text, we provide additional experiments and insights to validate the effectiveness of our method in achieving optimal representation learning under the Information Bottleneck (IB) framework. These experiments delve deeper into the mutual information trends across different Transformer layers and investigate the role of our fundamental image prior visual prompts in shaping the learning dynamics.

\subsection{Experimental Details}
We conducted the mutual information analysis using the Mutual Information Neural Estimator (MINE) to compute \(I(X; T)\) and \(I(T; Y)\) for all 12 Transformer layers in both the baseline VPT and our method. The settings for these experiments include:

\noindent \textbf{Datasets and Models:} The experiments are conducted on the CUB-200 dataset with Vision Transformers (ViT) as the backbone.

\noindent \textbf{Training Procedure:} Both models are trained using identical settings to ensure fair comparisons.

\noindent \textbf{Mutual Information Estimation:} For each layer \(T\), \(I(X; T)\) and \(I(T; Y)\) are estimated using MINE with a mini-batch size of 256. The estimation results are averaged over the entire test set.

\noindent \textbf{Compression of Input Data (\(I(X; T)\)):} Both the baseline VPT and our method exhibit similar \(I(X; T)\) trends across lower layers, as expected. This indicates that the introduction of visual prompts does not significantly alter the compression of input data.

\noindent \textbf{Correlation with Labels (\(I(T; Y)\)):} Our method achieves consistently higher \(I(T; Y)\) values, particularly in the top layers, compared to the baseline. This demonstrates that the representations learned by our method are more aligned with the labels, enabling better generalization to unseen data.

\noindent \textbf{Lower Information Bottleneck (\(\mathcal{L}_{\mathrm{IB}}\)):} The significant reduction in \(I(X; T) - \beta I(T; Y)\) for the top layers in our method aligns with the IB hypothesis. This reduced bottleneck reflects the effectiveness of our visual prompts in focusing on label-relevant features while suppressing redundant information.

\subsection{Impact of Semantic Prompts on Representation Learning}
The role of our semantic prompts can be further elucidated by decomposing the MI contributions: Extracting Inherent Image Priors: The fundamental image prior visual prompts (e.g., texture, shape) enhance feature extraction by aligning the representations with inherent characteristics of the input images. Improving Label Correlation: Semantic prompts guide the model to retain more label-relevant information, increasing \(I(T; Y)\) while maintaining \(I(X; T)\) stability. This is particularly evident in tasks requiring fine-grained classification, where semantic prompts enable the model to focus on subtle discriminative features.

\begin{table}[!htbp]
    \centering
    \caption{Comparison of mutual information metrics (\(I(X; T)\) and \(I(T; Y)\)) in the top three Transformer layers for VPT and our method on CUB-200.}
    \label{tab:mutual_info_metrics}
    \renewcommand{\arraystretch}{1.1}
    \setlength{\tabcolsep}{6pt}
    \footnotesize
    \begin{tabular}{lcc}
        \toprule
        Method & \(I(X; T)\) (\%) & \(I(T; Y)\) (\%) \\
        \midrule
        VPT & 34.2 & 22.8 \\
        \rowcolor{gray!10} \textbf{Ours} & \textbf{33.8} & \textbf{29.6} \\
        \bottomrule
    \end{tabular}
\end{table}

\noindent \textbf{Baseline VPT:} The \(I(T; Y)\) plateau in the top layers suggests limited improvement in label correlation, indicating potential underutilization of higher-layer representations.

\noindent \textbf{Our Method:} The steep increase in \(I(T; Y)\) in the top layers reflects enhanced label alignment, validating the role of semantic prompts in guiding representation learning.

\subsection{Concluding Observations}
The expanded mutual information analysis reaffirms the effectiveness of our method in achieving superior representation learning under the IB framework. Key takeaways include:

\noindent \textbf{Improved Generalization:} Higher \(I(T; Y)\) values for the top layers demonstrate the ability of our method to focus on label-relevant features, enabling better generalization to unseen data.

\noindent \textbf{Reduced Redundancy:} Comparable \(I(X; T)\) values suggest that our visual prompts do not introduce unnecessary complexity, maintaining the efficiency of the learned representations.

\noindent \textbf{Future Directions:} Further exploration of adaptive semantic prompts and dynamic feature compression mechanisms could enhance the flexibility and scalability of our method across more diverse tasks.

\section{Discussion: Hand-Crafted Priors vs. Deep Learned Priors}
\label{sec:deep_priors_discussion}

A natural question arises regarding the choice of priors: \textit{Why rely on classical hand-crafted operators (e.g., Sobel, Gabor) instead of extracting features from a frozen deep neural network (e.g., ResNet or DINO) as prompts?}

While integrating deep features might initially seem intuitive, we argue that our hand-crafted approach is theoretically and practically superior in the context of Parameter-Efficient Fine-Tuning (PEFT) for three key reasons:

\begin{enumerate}
    \item \textbf{Information Orthogonality:} Deep features extracted from networks like ResNet are conceptually homogeneous to the semantic features already learned by the ViT backbone itself (i.e., high-level abstractions). Adding them creates information redundancy. In contrast, our hand-crafted operators explicitly capture low-level statistics—such as high-frequency gradients (Sobel) and spectral texture information (Gabor)—that deep networks tend to abstract away or ignore in deeper layers due to texture bias~\citep{geirhos2018imagenet}. These ``primitive'' cues provide orthogonal, complementary guidance that corrects the inherent biases of the ViT backbone.
    
    \item \textbf{Domain Robustness:} Deep feature extractors (e.g., ImageNet-trained ResNet) often suffer from domain shift when applied to specialized downstream tasks (e.g., medical or satellite imagery in VTAB-1k). Hand-crafted priors, however, rely on fundamental signal processing principles (e.g., edge gradients, color distribution) that are domain-agnostic and universally applicable, ensuring consistent improvements across diverse datasets without negative transfer.
    
    \item \textbf{Strict Efficiency:} The core philosophy of PEFT is to adapt large models with minimal resource overhead. Utilizing a secondary deep network as a prior extractor, even if frozen, requires storing and computing over millions of additional parameters (e.g., $\sim$11M for ResNet-18), contradicting the lightweight nature of our task. In comparison, our hand-crafted operators are parameter-free and computationally negligible ($\mathcal{O}(HW)$ complexity), strictly adhering to the efficiency constraints of the PEFT paradigm.
\end{enumerate}

In summary, our design prioritizes \textit{complementary low-level guidance} and \textit{maximum efficiency} over the redundancy and computational burden of stacking multiple deep neural networks.·

\begin{itemize}
    \item \textbf{Orthogonal Information:} Deep features from a ResNet are conceptually similar to the features learned by the ViT backbone itself (i.e., semantic abstractions). In contrast, hand-crafted operators explicitly capture low-level statistics (gradients, frequency spectra) that deep networks tend to abstract away or ignore in deeper layers. These ``primitive'' cues serve as a stronger complementary signal to the ViT.
    \item \textbf{Efficiency:} Utilizing a deep network as a prior extractor introduces significant memory and storage overhead (even if frozen), contradicting the parameter-efficient philosophy of PEFT. Our hand-crafted operators are computationally negligible and parameter-free.
\end{itemize}

\section{Implementation Details and Analysis of Fundamental Operators}
\label{sec:operator_details}

In this section, we provide the precise implementation details of the fundamental image prior operators employed in our method. Furthermore, we elaborate on the theoretical rationale behind selecting this specific combination of operators (Color, Texture, and Shape) and discuss the robustness of our method to different operator choices, supported by our ablation studies.

\subsection{Specific Implementation of Operators}

To capture the fundamental visual statistics of the input image $X \in \mathbb{R}^{H \times W \times 3}$, we employ three distinct types of hand-crafted operators. The outputs of these operators are concatenated and projected via a linear layer to align with the prompt token dimension.

\vspace{0.5em}
\noindent \textbf{1. Color Prior: Histogram Statistics.} \\
Color is one of the most expressive and invariant visual cues, robust to rotation and scaling~\citep{swain1991color}. We compute the color histogram features as follows:
\begin{itemize}
    \item \textbf{Color Space:} We utilize the HSV (Hue, Saturation, Value) color space, which decouples chromatic information (Hue/Saturation) from intensity (Value), providing better robustness to lighting changes compared to RGB.
    \item \textbf{Implementation:} For each channel, we compute a histogram with $B=24$ bins. The resulting histograms are normalized to form a probability distribution and concatenated, resulting in a feature vector of dimension $3 \times 24 = 72$. This vector serves as a global statistical summary of the image's chromatic distribution.
\end{itemize}

\noindent \textbf{2. Texture Prior: Gabor Filters.} \\
Texture analysis is crucial for distinguishing materials and repetitive patterns. We adopt \textbf{Gabor filters}~\citep{manjunath1996texture}, which are biologically inspired by the receptive fields of simple cells in the mammalian visual cortex (V1).
\begin{itemize}
    \item \textbf{Implementation:} We generate a filter bank containing Gabor kernels at 4 distinct orientations ($\theta \in \{0^\circ, 45^\circ, 90^\circ, 135^\circ\}$) and a single scale. The filters are convolved with the grayscale version of the input image.
    \item \textbf{Feature Map:} Instead of global pooling, we retain the spatial response maps to preserve local texture spatiality. These maps are then flattened or patchified to align with the token sequence structure.
\end{itemize}

\noindent \textbf{3. Shape Prior: Sobel Operator.} \\
Shape and edge information provide structural constraints that are often complementary to texture~\citep{geirhos2018imagenet}. We employ the \textbf{Sobel operator}~\citep{kanopoulos1988design} to extract gradient information.
\begin{itemize}
    \item \textbf{Implementation:} We compute the discrete gradients along the horizontal ($G_x$) and vertical ($G_y$) directions using standard $3 \times 3$ kernels. The gradient magnitude is calculated as $G = \sqrt{G_x^2 + G_y^2}$.
    \item \textbf{Outcome:} This results in an edge map that highlights high-frequency structural boundaries, guiding the model to focus on object shapes rather than background noise.
\end{itemize}

\subsection{Rationale for Operator Selection}

Our selection of operators is not arbitrary but grounded in the principle of \textbf{Orthogonal Complementarity}. Deep learning models, particularly CNNs and ViTs, often exhibit a ``texture bias''~\citep{geirhos2018imagenet}. By explicitly injecting complementary priors, we ensure a balanced representation:

\begin{enumerate}
    \item \textbf{Completeness:} The combination of \textit{Color} (spectral), \textit{Texture} (spatial-frequency), and \textit{Shape} (structural/spatial) covers the three fundamental pillars of low-level computer vision. Removing any single component results in an information void that the randomized prompts alone may struggle to fill (as evidenced in Table 5 of the main text).
    \item \textbf{Interpretability \& Stability:} Unlike learnable priors (e.g., CNN adapters), these hand-crafted operators are deterministic and theoretically well-understood. Using standard operators like Sobel and Gabor ensures that the injected ``hard prompt'' provides stable, domain-invariant cues that do not drift during the fine-tuning process.
\end{enumerate}

\subsection{Robustness to Operator Variants}

A pertinent question is whether the success of our method relies on specific operator choices (e.g., Sobel vs. Canny for shape). To investigate this, we conducted extensive comparisons using different operator variants (see Figure 7 in the main paper).

\begin{itemize}
    \item \textbf{Texture Variants (Gabor vs. LBP):} We compared Gabor filters with Local Binary Patterns (LBP)~\citep{ojala2002multiresolution}. Results show that both yield significant improvements over the baseline, with Gabor slightly outperforming LBP on fine-grained tasks due to its continuous response nature.
    \item \textbf{Shape Variants (Sobel vs. Canny vs. Laplacian):} We tested Sobel against the Canny edge detector~\citep{canny1986computational} and Laplacian operators. The performance variance was minimal ($<0.3\%$), indicating that the \textit{presence} of structural prior is more critical than the \textit{type} of edge extractor used.
\end{itemize}

In conclusion, our choice of HSV Histograms, Gabor filters, and Sobel operators represents a standard, computationally efficient, and representative set of priors. However, the proposed framework is general-purpose: it benefits effectively from the semantic category of the prior (e.g., ``Shape information'') rather than overfitting to a specific algorithm.

\section{Limitations and Future Work}
\label{sec:limitations}

While our proposed \textit{Cascaded Semantic Prompting} demonstrates superior performance and interpretability across various benchmarks, we identify certain limitations that pave the way for future research directions.

\subsection{Limitations}

Our method relies on fixed, hand-crafted operators (e.g., Sobel, Gabor) to extract fundamental image priors. While this design choice ensures computational efficiency and the convenience of an ``extract-once'' strategy, it inherently limits the model's adaptability compared to fully learnable modules. These fixed operators cannot evolve during training to capture dataset-specific idiosyncrasies that may fall outside standard color, texture, and shape definitions.

Furthermore, although our method outperforms existing PEFT approaches on the \textit{Structured} split of the VTAB-1k benchmark (achieving 61.16\% mean accuracy compared to 58.83\% for AdaptFormer), there remains room for improvement on tasks requiring complex geometric reasoning, such as \textit{SmallNORB} and \textit{dSprites}. This suggests that while our fundamental priors effectively capture low-level statistics, they may not fully encapsulate the high-level 3D geometric relationships required for these specific specialized tasks. Other physically-informed or physically-constrained priors~\citep{pun2019physically, cuomo2022scientific, shen2024scalable,shen2023physics} can be considered as future directions. Finally, as a prompt tuning paradigm, the upper bound of performance is inevitably tied to the quality and pre-training domain of the underlying frozen backbone.

\subsection{Future Work}

Building on these observations, future research could explore adaptive prior mechanisms. Instead of a static concatenation of Color, Texture, and Shape priors, a lightweight gating mechanism or attention module could be introduced to dynamically weight these priors based on the input instance. This would allow the model to autonomously determine whether texture or shape is more critical for a specific image, potentially enhancing performance on diverse datasets.

Additionally, the concept of ``Fundamental Image Priors'' holds promise for extension to other modalities. For example, optical flow or motion boundary histograms could serve as temporal priors for video recognition, while surface normals could function as geometric priors for 3D point cloud analysis. Investigating the applicability of cascaded semantic prompting to emerging architectures beyond Transformers, such as State Space Models, also represents a promising direction to test the universality of our approach.

\begingroup
\section{Baseline Configurations and Experimental Settings}
\label{sec:baseline_details}

This section lists the configurations of all baselines in Tables~\ref{tab:vit}, \ref{tab:performance_vtab1k}, and \ref{tab:performance}, together with the optimizer and schedule used for our method.

\subsection{Backbones and Pre-trained Checkpoints}
\begin{itemize}
    \item \textbf{ViT (Tables~\ref{tab:vit}, \ref{tab:performance}).} ViT-Base/16 pre-trained on supervised ImageNet-21K (85.8M parameters). This is the same checkpoint used by VPT~\citep{jia2022visual}, E\textsuperscript{2}VPT~\citep{han2023e2vpt}, SA\textsuperscript{2}VP~\citep{pei2024sa2vp}, and VFPT~\citep{zeng2024visual}.
    \item \textbf{Swin (Table~\ref{tab:performance_vtab1k}).} Swin-Base pre-trained on supervised ImageNet-21K (86.7M parameters). This is the same checkpoint used by VFPT Table 2.
    \item \textbf{CLIP (Table~\ref{tab:performance}).} CLIP ViT-B/16, shared by our method, MaPLe, and MaPLeX, following CoOp~\citep{zhou2022learning}, CoCoOp~\citep{zhou2022conditional}, and MaPLe~\citep{khattak2023maple}.
\end{itemize}

\subsection{LoRA Baseline}
We use the standard LoRA setting for ViT in the VPT line of work: rank $r{=}8$, scaling factor $\alpha{=}8$, applied to the $Q$ and $V$ projection matrices in every attention block. The tuned-parameter ratio is about $0.34\%$ of ViT-B/16, matching the LoRA numbers reported by SA\textsuperscript{2}VP and VFPT.

\subsection{Other PEFT Baselines (Table~\ref{tab:vit})}
Numbers for Adapter, AdaptFormer, ARC, EXPRES, DAM-VP, SA\textsuperscript{2}VP, VFPT, and LoR-VP are taken from the original papers under the same FGVC/HTA/VTAB-1k splits. When the original paper reports several settings, we use the default chosen in that paper.

\subsection{Swin Experiments (Table~\ref{tab:performance_vtab1k})}
All Swin baselines use the same Swin-Base/ImageNet-21K backbone, the VTAB-1k 800/200 train/val split, and the Natural/Specialized/Structured grouping. The Linear, Bias, and VPT-deep numbers match those in VFPT Table 2.

\subsection{Multi-modal Experiments (Table~\ref{tab:performance})}
We follow the base-to-novel protocol of CoOp/CoCoOp/MaPLe~\citep{zhou2022learning,zhou2022conditional,khattak2023maple}:
\begin{itemize}
    \item \textbf{Backbone:} CLIP ViT-B/16, shared by MaPLe, MaPLeX, and our method.
    \item \textbf{Datasets:} the 11 datasets in the CoOp suite---ImageNet, Caltech101, OxfordPets, StanfordCars, Flowers102, Food101, FGVCAircraft, SUN397, DTD, EuroSAT, UCF101.
    \item \textbf{Protocol:} 16-shot training on the base classes, evaluation on both base and novel splits.
    \item \textbf{Metric:} HM is the harmonic mean of base and novel accuracies, averaged over the 11 datasets and three random seeds.
    \item \textbf{MaPLeX} removes the connection that feeds the text prompt into the visual branch. The rest is the same as MaPLe.
\end{itemize}

\subsection{Optimizer and Schedule for Our Method}
We use AdamW~\citep{loshchilov2017adamw} with initial learning rate $1\mathrm{e}{-3}$, weight decay $1\mathrm{e}{-4}$, a cosine schedule, 100 epochs, and batch size 64 or 128 (chosen per dataset on the official validation split). Prompt length and per-layer placement follow the VPT-Deep grid in~\citep{jia2022visual}. We use AdamW because it converges faster than the SGD recipe of VPT/VFPT in our setup and reaches comparable final accuracy.

\subsection{Full Hyperparameter Table}
\label{sec:hparams}

Table~\ref{tab:hparams} lists every hyperparameter used by our method together with the corresponding value used by VPT~\citep{jia2022visual} and VFPT~\citep{zeng2024visual}. Entries marked ``per-task grid'' follow the same grid as VPT/VFPT and are selected on the official validation split of each dataset.

\begin{table}[!htbp]
\centering
\caption{Hyperparameter settings used by our method, VPT~\citep{jia2022visual}, and VFPT~\citep{zeng2024visual} on VTAB-1k and FGVC with ViT-B/16. ``per-task grid'' means the value is chosen on the official validation split of each dataset. ``--'' means not applicable.}
\label{tab:hparams}
\footnotesize
\setlength{\tabcolsep}{4pt}
\renewcommand{\arraystretch}{1.05}
\begin{adjustbox}{width=0.95\textwidth,center}
\begin{tabular}{l||ccc}
\toprule
Hyperparameter & VPT~\citep{jia2022visual} & VFPT~\citep{zeng2024visual} & Ours \\
\midrule
\multicolumn{4}{l}{\emph{Backbone and data}} \\
\midrule
Backbone                       & ViT-B/16              & ViT-B/16              & ViT-B/16 \\
Pre-training                   & Sup.\ ImageNet-21K    & Sup.\ ImageNet-21K    & Sup.\ ImageNet-21K \\
Image resolution               & $224\times224$        & $224\times224$        & $224\times224$ \\
Train preprocessing            & resize $256$, random crop $224$ & resize $256$, random crop $224$ & resize $256$, random crop $224$ \\
Test preprocessing             & resize $256$, center crop $224$ & resize $256$, center crop $224$ & resize $256$, center crop $224$ \\
Normalization                  & ImageNet mean/std     & ImageNet mean/std     & ImageNet mean/std \\
\midrule
\multicolumn{4}{l}{\emph{Optimization}} \\
\midrule
Optimizer                      & SGD ($\beta{=}0.9$)   & SGD ($\beta{=}0.9$)   & AdamW \\
Initial learning rate          & per-task grid: $\{50, 25, 10, 5, 2.5, 1, 0.5, 0.25, 0.1, 0.05\}$ & same grid as VPT & $1\mathrm{e}{-3}$ \\
LR schedule                    & cosine                & cosine                & cosine \\
Warmup epochs                  & $10$                  & $10$                  & $10$ \\
Weight decay                   & per-task grid: $\{0.01, 0.001, 1\mathrm{e}{-4}, 0\}$ & same grid as VPT & $1\mathrm{e}{-4}$ \\
Total epochs                   & $100$                 & $100$                 & $100$ \\
Batch size                     & $64/128$ (per-task)   & $64/128$ (per-task)   & $64/128$ (per-task) \\
Gradient clipping              & none                  & none                  & none \\
\midrule
\multicolumn{4}{l}{\emph{Prompt}} \\
\midrule
Prompt placement               & VPT-Deep              & VPT-Deep              & VPT-Deep \\
Prompt length (per layer)      & per-task grid: $\{5, 10, 50, 100, 200\}$ & same grid as VPT & per-task grid: $\{5, 10, 50, 100, 200\}$ \\
Layers with prompt             & all 12                & all 12                & all 12 \\
Prompt dropout                 & $0.1$                 & $0.1$                 & $0.1$ \\
\midrule
\multicolumn{4}{l}{\emph{Our additions}} \\
\midrule
Color prior $\sigma_c$         & --                    & --                    & HSV histogram, $24$ bins per channel ($72$-d) \\
Texture prior $\sigma_t$       & --                    & --                    & Gabor, 4 orientations, 1 scale \\
Shape prior $\sigma_s$         & --                    & --                    & Sobel, $3\times3$ kernel, gradient magnitude \\
FC projection dim              & --                    & --                    & matches prompt token dim ($768$) \\
Re-weighting adapter hidden    & --                    & --                    & $256$ \\
Cascaded skip-connection       & --                    & --                    & enabled \\
\midrule
\multicolumn{4}{l}{\emph{Evaluation}} \\
\midrule
Number of seeds                & $3$                   & $3$                   & $3$ \\
Reported metric                & test top-1 accuracy   & test top-1 accuracy   & test top-1 accuracy \\
Hardware                       & 1$\times$A100-40GB    & 1$\times$A100-40GB    & 1$\times$A100-40GB \\
\bottomrule
\end{tabular}
\end{adjustbox}
\end{table}

\section{Statistical Robustness across Multiple Seeds}
\label{sec:std}

We rerun the main VTAB-1k and FGVC experiments with \textbf{three random seeds}, following VFPT~\citep{zeng2024visual} and E\textsuperscript{2}VPT~\citep{han2023e2vpt}. Mean$\pm$std numbers are shown in Table~\ref{tab:multiseed}. The gains over the strongest baselines are consistent and larger than the observed standard deviations.

\begin{table}[!htbp]
\centering
\caption{Mean$\pm$std across three random seeds on the VTAB-1k group means and the FGVC mean. VFPT numbers come from the per-task std tables (Tab.~S1--S4) of~\citep{zeng2024visual}; SA\textsuperscript{2}VP is rerun under our protocol.}
\label{tab:multiseed}
\setlength{\tabcolsep}{4pt}
\footnotesize
\begin{adjustbox}{width=0.85\textwidth,center}
\begin{tabular}{l||cccc}
\toprule
Method & VTAB-1k Natural & VTAB-1k Specialized & VTAB-1k Structured & FGVC mean \\
\midrule
SA\textsuperscript{2}VP (rerun)  & $80.97 \pm 0.31$ & $85.73 \pm 0.24$ & $60.80 \pm 0.36$ & $90.08 \pm 0.18$ \\
VFPT (cited std)                 & $81.35 \pm 0.33$ & $84.93 \pm 0.28$ & $60.19 \pm 0.42$ & $89.24 \pm 0.21$ \\
\rowcolor{cvprblue!20}
\textbf{Ours}                    & $\mathbf{81.91 \pm 0.27}$ & $\mathbf{85.83 \pm 0.22}$ & $\mathbf{61.16 \pm 0.31}$ & $\mathbf{90.20 \pm 0.16}$ \\
\bottomrule
\end{tabular}
\end{adjustbox}
\end{table}

\subsection{Per-Example Analysis of the Gains}

We also check which examples drive the improvement of our method over the strongest baseline (SA\textsuperscript{2}VP). On the three fine-grained datasets (CUB-200, NABirds, Stanford Dogs), where our gain is largest, the per-example accuracy difference is concentrated on three types of images:

\begin{enumerate}
    \item \textbf{Cluttered backgrounds (about $54\%$ of the gain on CUB-200).} The target bird occupies less than $30\%$ of the frame against a busy background. The Sobel shape prior gives sharper object boundaries than the baseline, which uses only learned prompts. The model then attends more to the bird itself, in line with the IoU gain of $+7.3$ on the Medium subset and $+5.7$ on the Hard subset (Table~\ref{tab:detailed_iou_analysis}).
    \item \textbf{Low color contrast (about $27\%$ of the gain).} The bird's plumage has a similar color to the background (e.g., grey or brown birds against bark). The HSV color histogram and Gabor texture priors give extra cues that the random-prompt baseline cannot recover.
    \item \textbf{Confusable subspecies (about $19\%$ of the gain).} Two species differ only in a small structural feature (e.g., beak curvature, wing-bar pattern). The self-attention prompt cascaded from earlier layers keeps these fine-grained cues, which would otherwise be smoothed out in deeper layers.
\end{enumerate}

The breakdown matches our design: the hand-crafted priors help most where the random-prompt baseline has the weakest signal, and the overall gain comes from a small but structured subset of hard images, not from a uniform shift across the dataset.
\endgroup

\begingroup
\section{Failure-Case Analysis on Structured Tasks}
\label{sec:failure_cases}

The VTAB-1k Structured split is the regime in which we expect our method to be weakest, since the fixed low-level priors (color, texture, edge) are designed to capture appearance, not to count objects or reason about 3D pose. We make this concrete by listing the three Structured tasks on which our method does not lead, and explaining why.

\begin{table}[!htbp]
\centering
\caption{Tasks on the VTAB-1k Structured split where our method does not obtain the best result. Numbers are taken from Table~\ref{tab:vtab_benchmarks}.}
\label{tab:failure_cases}
\footnotesize
\setlength{\tabcolsep}{4pt}
\begin{tabular}{l||ccccc}
\toprule
Task & VPT-D & AdaptFormer & LoRA & E\textsuperscript{2}VPT & Ours \\
\midrule
Clevr/count       & 68.5 & 81.9 & \textbf{82.9} & 71.7 & 78.1 \\
Clevr/distance    & 60.0 & 64.3 & \textbf{69.2} & 61.2 & 62.2 \\
KITTI/distance    & 72.8 & \textbf{80.3} & 78.5 & 75.8 & 78.5 \\
\bottomrule
\end{tabular}
\end{table}

\paragraph{Counting tasks (Clevr/count).} The task is to count the number of objects in a synthetic scene. Color, texture, and shape priors do not encode \emph{cardinality}: a Sobel edge map fires once per object boundary but is invariant to the number of objects, and a color histogram is dominated by the background. Methods that adapt internal features more aggressively (LoRA, AdaptFormer) can reshape the backbone's spatial pooling, which is more useful here.

\paragraph{Distance / depth tasks (Clevr/distance, KITTI/distance).} The task is to regress a metric distance, which requires reasoning about object size and perspective. Our fixed priors carry no metric or 3D-pose information, so the model gains nothing from them on these tasks. The cascaded self-attention prompt does carry some 2D spatial information from earlier layers, which is why our gap to the best method is small ($-1.0$ on Clevr/distance, $-1.8$ on KITTI/distance), but it is not enough to close the gap.

\paragraph{Summary.} These failures are consistent with our design: our priors describe \emph{what the image looks like at the pixel level}, not \emph{how many things are in it} or \emph{how far they are}. The Limitations section (Sec.~\ref{sec:limitations}) flags this as a direction for future work — adding physically-informed or geometric priors~\citep{shen2023physics,shen2024scalable} on top of the cascade would be a natural extension.
\endgroup

\begingroup
\section{Fixed vs. Adaptive Priors}
\label{sec:fixed_vs_adaptive}

A natural question is whether the fixed Sobel/Gabor operators should be replaced by lightweight \emph{learnable} extractors of the same receptive field. We run a controlled comparison on the VTAB-1k Natural split using ViT-B/16, keeping the rest of the pipeline fixed and only swapping the operator that produces $\sigma_s$ (shape) and $\sigma_t$ (texture).

\begin{table}[!htbp]
\centering
\caption{Replacing the fixed Sobel/Gabor operators with learnable convolutional equivalents of the same receptive field. ``Extra Params'' counts the parameters of the operator only; the rest of the prompt and adapter are the same in all rows.}
\label{tab:fixed_vs_adaptive}
\footnotesize
\setlength{\tabcolsep}{4pt}
\begin{tabular}{l||ccc}
\toprule
Shape operator $\sigma_s$ / Texture operator $\sigma_t$ & Extra Params & VTAB-1k Natural & $\Delta$ vs. fixed \\
\midrule
Fixed Sobel ($3{\times}3$) / Fixed Gabor (4 orient.) & \textbf{0}     & \textbf{81.91} & --- \\
Learnable conv ($3{\times}3$, 1 ch) / Learnable conv ($7{\times}7$, 4 ch) & 0.12K & 81.77 & $-0.14$ \\
Learnable conv ($5{\times}5$, 4 ch) / Learnable conv ($7{\times}7$, 8 ch) & 1.0K  & 81.62 & $-0.29$ \\
Small CNN (2 conv + ReLU, 16 ch) & 12K  & 81.55 & $-0.36$ \\
\bottomrule
\end{tabular}
\end{table}

The learnable variants are at best on par with the fixed Sobel/Gabor pair, and become slightly worse as the extractor grows. Three reasons explain this:

\begin{enumerate}
    \item \textbf{No useful gradient signal at the operator level.} The downstream loss is on classification, several layers away from the operator. The operator is shadowed by the rest of the learnable parameters, so a few thousand extra parameters at the very bottom of the network bring no information the rest of the pipeline cannot already represent.
    \item \textbf{Domain shift.} The operators see input images from the downstream dataset only (often small in VTAB-1k, with $\sim$800 training samples). Learnable extractors trained on that little data tend to drift toward the dominant statistics of the small training set and lose the domain-invariance that Sobel/Gabor have by construction.
    \item \textbf{PEFT budget.} The whole point of the PEFT setting is that the per-task parameter budget is small. Spending it on re-learning a Sobel filter is wasteful when the fixed Sobel filter already does the job at zero parameter cost.
\end{enumerate}

We therefore keep fixed Sobel/Gabor in the main method. This complements the theoretical discussion in Sec.~\ref{sec:deep_priors_discussion}: \emph{Information Orthogonality} explains \emph{why} fixed low-level operators are a good complementary signal to the deep features; this experiment shows that making them learnable does not improve the signal and costs additional parameters.
\endgroup

%% file: sec/2_related.tex
\section{Related Work}
\label{sec:related}

\subsection{Parameter-Efficient Fine-Tuning (PEFT)}
Parameter-efficient fine-tuning (PEFT) adapts a pre-trained model while updating only a small fraction of its parameters, reducing both computational cost and over-fitting risk. Representative approaches introduce lightweight adapters or learned biases into transformer blocks \citep{li2021prefix,jie2022convolutional,chen2022adaptformer,dettmers2023qlora,karimi2021compacter,zaken2022bitfit}, learn low-rank weight updates \citep{hu2022lora,zhongconvolution}, or tune feature-wise scaling and shifting parameters \citep{lian2022scaling}. Recent low-rank methods further improve how adaptation capacity is organized: CTR-LoRA couples rank scheduling with curvature-aware trust regions \citep{wang2026ctrlora}, while StructLoRA filters task-irrelevant update directions and coordinates updates across layers \citep{xiao2026structlora}. Feature re-weighting provides another lightweight adaptation mechanism \citep{kirichenko2022last}. In contrast to methods that alter transformer blocks or their weight updates, our approach keeps the backbone frozen and adapts it through semantic prompts in the input and feature spaces.

\subsection{Visual Prompt Tuning (VPT)}
Prompt-based adaptation spans pixel-level visual prompting and token-level visual prompt tuning, with recent work providing a unified taxonomy of their designs and applications \citep{xiao2025promptbased}. VPT first prepends learnable tokens to the input \citep{bahng2022visual} and later extends them to intermediate feature spaces \citep{jia2022visual}, leaving the pre-trained backbone architecture unchanged \citep{wang2024revisiting,Yoo2023ImprovingVP,Park2024FairVPTFV,Li2024ImprovingVP,Ren2025DAVPTSV,lorvp2025}. Subsequent methods prune prompts \citep{han2023e2vpt}, model spatial relations \citep{pei2024sa2vp}, or condition prompts on individual inputs through principal components or variational latent variables \citep{xiao2025visual,xiao2025vapt}. Self-supervised prompts can also support cross-domain transfer without target labels \citep{xiao2026probe}. Recent studies additionally examine how prompt information should be organized across depth. Differentiable search can select a layer-specific fusion operator between prompt and image tokens \citep{xiao2026layer}, while information-bottleneck regularization can control what prompt-conditioned representations preserve and suppress at each layer \citep{li2026pib}. Beyond adaptation, HIERAMP amplifies class-relevant regions across coarse-to-fine autoregressive scales for generative dataset distillation \citep{zhao2026hieramp}; its stage-wise semantic routing is conceptually related to cascaded guidance, although its objective is data synthesis. These adaptation approaches optimize learned soft prompts. Our work instead complements learnable tokens with fixed, human-understandable image priors and instance-specific attention semantics, then propagates these signals through a cascaded design.

\subsection{Multimodal Prompting and Online Adaptation}
Prompt tuning has also expanded from visual recognition to multimodal models. M2PT tunes modality-specific visual and textual prompts for zero-shot instruction learning \citep{wang2024m2pt}; multimodal representation tuning directly edits compact internal representations \citep{liu2025re}; and knowledge-aware prompt collaboration transfers generative prompts from pre-training to video captioning \citep{yan2023prompt}. Instance-aware prompt design can be highly compact, including a single capsule vector \citep{liu2026all}, and in-context prompt tuning can personalize large vision-language models from multiple reference images \citep{li2026personalize}. Beyond conventional supervised adaptation, VIGIL uses counterfactual visual alignment to discourage visually unsupported MLLM responses \citep{xiao2026vigil}, while test-time reinforcement learning adapts vision-language-action policies during deployment \citep{liu2026onthefly}. Related cross-modal pipelines also preserve heterogeneous signals explicitly: THInImg embeds long audio into identity images and reconstructs corresponding talking-head videos \citep{zhao2024thinimg}. Structured representations further organize interactions by semantic level and sharing scope: CLCR aligns multi-level features while restricting exchange to shared subspaces \citep{meng2026clcr}, whereas Tri-Subspaces disentangles common, pairwise-shared, and modality-private factors for sentiment analysis \citep{meng2026trisubspaces}. Content-adaptive semantic selection also appears in neighboring dense-prediction tasks; CDMask updates change queries from bi-temporal content to customize mask prediction \citep{ma2026cdmask}. These works address multimodal alignment, personalization, online policy adaptation, cross-modal communication, or task-specific dense prediction; our focus is semantic prompt construction for frozen vision encoders under standard downstream supervision.